\documentclass[twocolumn,10pt]{asme2ej}
\usepackage[colorinlistoftodos]{todonotes}

\usepackage{fancyhdr,xcolor}
\usepackage{graphicx} 
\usepackage{amsmath}
\usepackage{hyperref}
\usepackage{multirow}
\usepackage{caption}
\usepackage{subcaption}
\usepackage{makecell}
\usepackage{siunitx,booktabs,xcolor}
\title{GANTL: Towards Practical and Real-Time Topology Optimization with Conditional GANs and Transfer Learning}

\author{Mohammad Mahdi Behzadi 
    \affiliation{Computational Design Laboratory\\
	Department of Mechanical Engineering\\
	University of Connecticut\\
    Email: mohammad.behzadi@uconn.edu
    }	
}

\author{Horea T. Ilie\c{s}\thanks{Corresponding author.} 
    \affiliation{Computational Design Laboratory\\
	Department of Mechanical Engineering\\
	University of Connecticut\\
    Email: horea.ilies@uconn.edu
    }
}

\begin{document}

\maketitle    

\begin{abstract}
{\it Many machine learning methods have been recently developed to circumvent the high computational cost of the gradient-based topology optimization. These methods typically require extensive and costly datasets for training, have a difficult time generalizing to unseen boundary and loading conditions and to new domains, and do not take into consideration topological constraints of the predictions, which produces predictions with inconsistent topologies. 

We present a deep learning method based on generative adversarial networks for generative design exploration. The proposed method combines the generative power of conditional GANs with the knowledge transfer capabilities of transfer learning methods to predict  optimal topologies for unseen boundary conditions. We also show that the knowledge transfer capabilities embedded in the design of the proposed algorithm significantly reduces the size of the training dataset compared to the traditional deep learning neural or adversarial networks.  Moreover,  we formulate a topological loss function based on the bottleneck distance obtained from the persistent diagram of the structures and demonstrate a significant improvement in the topological connectivity of the predicted structures. We use numerous examples to explore the efficiency and accuracy of the proposed approach for both seen and unseen boundary conditions in 2D.
}
\end{abstract}


\section{Introduction}

Topology optimization (TO) is an iterative procedure that outputs the optimum material distribution of a structure within a specified domain subject to prescribed boundary conditions and external loads \cite{bendsoe2013topology,sigmund2013topology}. Typically, gradient-based techniques combined with model parameter continuation are used to promote convergence to optimal solutions. The process often requires hundreds of iterations, each involving complete finite element solutions,  which are time consuming \cite{amir2011reducing}. For example, the SIMP solver presented in \cite{andreassen2011efficient} needs 1.85s per iteration and hundreds of iterations to solve a simple MMB beam with resolution (300 x 100).

Over the past few years, various machine learning approaches have been proposed to improve the computational cost of gradient-based topology optimization, but the existing methods have significant limitations, which limit their utility in practice. Without a doubt, a practical machine-learning based topology optimization tool that could be used to explore the associated design space needs to: (1) be trainable with a relatively small training data set; (2) have a strong generalization ability to boundary and loading conditions as well as design domains that are not part of the training data set; and (3) enforce the topological connectivity of the predicted structure. 

For now, the proposed machine learning methods that make pixel-wise predictions of the optimum structures need extensive training datasets, and especially those with generative characteristics, and the generation of these large datasets is very expensive from a computational standpoint. Moreover, only a few of the proposed algorithms have been shown to handle boundary conditions that are not part of the training data, but in all these cases, the unseen boundary conditions are very close to those used in the training data. At the same time, and without exception, the structural predictions made by all these methods have not only poor but also inconsistent topologies compared to the ground truth. The latter limitation, which increases in severity with the increase in resolution, is due to the fact that none of these methods takes into consideration the topology of the prediction.  

In our recent paper  \cite{behzadi2021real}, we introduced a transfer learning approach based on Convolutional Neural Networks (CNN) that was capable of handling high-resolution 2D \& 3D design domains of various shapes and topologies and supporting real-time design space explorations. We showed that the knowledge transfer capabilities of that network significantly reduced the size of the training dataset compared to the traditional deep learning neural networks. Consequently, we showed that the proposed architecture was capable of handling boundary conditions that were unseen to the source network by only fine-tuning the corresponding target network with a much smaller training data set. While the method demonstrated the knowledge transfer capabilities of  transfer learning applied to topology optimization, it requires fine-tuning of the target network for unseen domains and boundary conditions. 

Generative Adversarial Networks (GAN) are one of the most promising developments in machine learning because they have the ability to learn low and high-level patterns in datasets and use them to generate new data that bears a (statistical) resemblance to the original dataset. As detailed in section \ref{background:sec}, GANs have been used to generate realistic but artificially generated images, and these same generative properties prompted GANs to be used in making predictions of optimal topologies. However, standard GANs need very large training data sets to be able to generate the new images, and their training can be notoriously difficult due to mode collapses and diminished gradients.

In this paper, we combine the generative power of conditional GANs with the knowledge transfer capabilities of transfer learning methods to predict optimal topologies for unseen boundary conditions. We show that the knowledge transfer capabilities embedded in the design of the proposed algorithm significantly reduces the size of the training dataset compared to the traditional deep learning neural or adversarial networks.  In fact, we show that using the same amount of data as in \cite{behzadi2021real}, the proposed GANTL can generate the structures with new boundary conditions that are not included in the target model training data. Moreover,  we formulate a topological loss function based on the bottleneck distance obtained from the persistent diagram of the structures and demonstrate a significant improvement in the topological connectivity of the predicted structures. We use numerous examples to explore the efficiency and accuracy of the proposed approach for both seen and unseen boundary conditions. To the best of our knowledge, this is the first time that the topological loss function is combined with GAN to tackle the connectivity issue of the predicted optimum structures.

\section{Background}\label{background:sec}

The significant computational cost of the traditional gradient-based TO prompted the investigation of various alternatives to the highly iterative process. For instance, parallelized algorithms run on high-performance computing platforms with a multigrid preconditioner was discussed in \cite{traff2021topology} to optimize `ultra large' scale shell models comprised of over 11M elements in 17 hours using 800 compute cores. The parallelized level set method \cite{liu2019fully} solves a 96 × 48 × 24 cantilever beam in 45 minutes on a desktop computer with 4  Intel i7-6700 CPU cores and 16 GB memory. Moreover, the work described in \cite{aage2015topology} employs the Portable and Extendable Toolkit for Scientific Computing (PETSc) to produce the optimal topology starting from a cantilever beam with 480 × 240 × 240 elements in 14 hours using 144 CPU cores.

The high computational cost of the gradient-based TO algorithms prompted the development of a number of machine learning methods that can make predictions almost instantaneously once trained. Recently proposed CNN-based algorithms can predict the optimum structure for low resolution 2D domains (50 x 50, and 40 x 40, respectively) \cite{abueidda2020topology,sosnovik2019neural} and  low resolution 3D domain (24 x 12 x 12)  \cite{banga20183d}. 

Several recent papers in the realm of topology optimization were inspired by the ability of GANs to synthesize images of human faces that don't belong to a real person \cite{karras2017progressive}, which learn the patterns and the distribution of the input data - see also section \ref{sectionGAN}.  Despite this, many TO approaches that use GANs do not explore this important aspect.  For example, conditional Wasserstein generative adversarial networks (CWGAN), which minimize an approximation of the Earth Mover's distance \cite{arjovsky2017wasserstein}, were used in \cite{rawat2019novel}  for 2D TO.  That method was trained on 3,024 cantilever beams (120 x 120) with design variables that include the volume fraction, penalty factor, and radius of the filter. They concatenate their design variables with the noise vector and feed them to their CWGAN to generate the cantilever beam that corresponds to the design variables. A two-stage hierarchical prediction–refinement GAN-based framework \cite{li2019non}  is used to predict the low resolution near-optimal structure and its corresponding refined structure in high resolution. Their training data set contains 9,900 pairs of low-resolution (40 x 40) and high-resolution (160 x 160) data, and the method achieves a 9.1 \% MSE for high resolution predictions. Similarly,  \cite{yu2019deep} describes a two-stage refinement cGAN to predict the high resolution optimum structure, which is trained with 64,000 low (32 × 32) resolution and their high (128 × 128) resolution counterparts.

However, all these algorithms require large datasets for training, lack generalization abilities, and do not have a mechanism to enforce the topology of their predictions. 

Two recent papers do provide some evidence of generalizability. Specifically, the method described in \cite{zhang2019deep} uses 64,000 cantilever beam to train a deep CNN to predict the optimum structure for simple low-resolution (40 x 80) 2D domains. Their network uses the displacement and the strain field as inputs, and the authors showed that their method could generate a solution for the simply supported and the two-span continuous boundary conditions that are not included in the training data set. Moreover,  a
conditional GAN, named TopologyGAN, is described in \cite{nie2021topologygan}, which used  49,078 2D beams (64 x 128) with 42 different displacement boundary conditions to train for their model.  Their results display a nearly three times reduction in the MSE on test problems involving four new boundary conditions that are not part of the training data set comparing to the baseline cGAN. Although both methods have shown some success in making predictions for domains with unseen boundary conditions, they do require large training datasets, and they do not consider the topology of the predictions that they make, often producing severely branched (tree-like) predictions.

To date, the only known method to produce well-connected predictions involves variants of the gradient-based TO methods, which trade computational efficiency for an improved connectedness. For instance, \cite{chandrasekhar2020tounn}, and \cite{chandrasekhar2021multi} proposed an iterative topology optimization method using neural networks for single material (TOuNN) and multi-material structures (MM-TOuNN) for direct execution of TO. In this work, the NN weights and bias are used to parameterize the density function,  thus disconnecting the density from the finite element mesh and obtain a crisp and differentiable material interface and boundary. The NN is used to predict the density values at each iteration, and part of the associated sensitivity analysis was computed via backpropagation. Importantly, the iterations performed by this approach are still driven by FEA. Clearly, this variation of the gradient-based TO is flexible in terms of the domains and the boundary conditions it can tackle, since it continues to be driven by FEA, but remains computationally expensive and seems to lose some of SIMP's ability to reproduce the solutions generated by SIMP.  For example, some examples from \cite{chandrasekhar2020tounn} require twice as much time per iteration compared to SIMP \cite{andreassen2011efficient}. 

In this paper, we take a different approach and integrate notions from persistence homology \cite{edelsbrunner2008persistent, cohen2007stability} to consider the connectedness of the predicted structure. As argued above, such topological properties represent one of the key criteria for making these machine learning algorithms practical.  However, generating predictions with topological guarantees is certainly not specific to the field of topology optimization. In fact, the image processing community has been interested in techniques that capture the topological properties of image predictions. For example, \cite{mosinska2018beyond} uses feature maps from pre-trained very deep convolutional networks, dubbed as VGG, obtained for the ground truth and the predicted image to construct a topology-aware loss function. They showed that by minimizing this loss function, one can significantly improve the topological features of the predicted image. Another way to improve the topological feature of the predicted images is using the persistent homology to capture the topological information of the images in order to construct the topological loss function. The work described in \cite{hu2019topology} used the modified Wasserstein distance between the persistent diagram of the ground truth and the prediction to construct the loss function. This method can achieve much better performance in terms of the Betti number\footnote{A reasonably accessible description of homology groups and Betti numbers can be found in \cite{munkres2018elements}.} error between the prediction and ground truth than any other prior method.

\subsection{Contributions and Outline}
The data-driven topology optimization method proposed in this paper combines the generative power of conditional GANs with the knowledge transfer capabilities of transfer learning methods. We show that this combination results in a practical TO tool based for design space exploration that: (1) requires a much smaller training data set than any other data-driven TO method; (2) has appealing generalization abilities for unseen boundary conditions; (3) displays a significant improvement in the topological connectivity of the predicted structures due to the new loss function that uses key concepts from persistence homology.

The rest of the paper is organized as follows. Section \ref{formulation:sec} contains the preliminaries, including formulation, GANTL architecture, data generation, and metrics used for model evaluation. We demonstrate the generality and flexibility of our approach by exploring various 2D examples with different resolutions, boundary conditions, and design spaces, including boundary conditions that are \textit{unseen} the network. These examples show that the proposed method supports real-time generative design space explorations for both \textit{seen} and \textit{unseen} boundary condition and the topological improvements of our predictions due to the novel topology-aware loss function. Finally, Section \ref{conclusions:sec} summarizes the key advantages and limitations of the proposed method and its potential applications.

\section{Problem Formulation}\label{formulation:sec}

\subsection{Minimum Compliance TO problem}
Topology optimization seeks the material distribution $\rho(\mathbf{x})$ that minimizes an objective function $f(\Omega, \rho)$ that is subjected to various constraints $g_i \leq 0$. A common objective function in structural TO is the global structural compliance \cite{sigmund200199,rozvany2001stress,bendsoe2007topology}, but local objective functions, such as those depending on the von Mises stresses, have been proposed \cite{yang1996stress,le2010stress}. The minimization is usually carried out by solving a finite element analysis.  The well known SIMP approach \cite{bendsoe1989optimal} associates each element with a density variable $\rho_{e} \in [0,1]$, where $0$ corresponds to an empty element, and $1$ to an element completely filled with material. The corresponding optimization problem can be formulated as 

\begin{eqnarray}\label{eq.1}
   min &:& f(\Omega, \mathbf{\rho}) = \mathbf{U^{t}KU} = \sum E_{e}(\rho_{e})\mathbf{u}_{e}^{t}\mathbf{k}^0_{e}\mathbf{u}_{e} \\
    s.t  &:& \mathbf{KU} = \mathbf{F} \\
    		 & & \sum \rho_{e}v_{e} \leq V_{max} \\  
         & &  0\leq \rho \leq 1
\end{eqnarray}
where $\Omega$ is the design domain; $\mathbf{\rho}$ is the density vector; $\mathbf{U}$ and $\mathbf{F}$ are the global displacement and force vectors, respectively; $\mathbf{K}$ is the global stiffness matrix; $\mathbf{u}_{e}$ is the element displacement vector; $\mathbf{k}^0_{e}$ is the element stiffness matrix for an element with unit Young's modulus; $v_{e}$ and $\rho_{e}$ are the element volume and density of element $e$, respectively; and $V_{max}$ is the volume upper bound. $E_{e}(\rho_{e})$ is the element's Young's modulus determined by the element density $\rho_{e}$:
\begin{equation}
    E_{e}(\rho_{e}) = E_{min} + \rho_{e}^{p}(E_{0}-E_{min}), \hspace{1cm}  \rho_{e} \in[0,1] \label{eq.2}
\end{equation}
In equation (\ref{eq.2}), $E_{0}$ represents the stiffness of the material, $E_{min}$ is a very small number, and $p$ is a penalty factor \cite{sigmund2013topology,sigmund2007morphology}.
 
\subsection{Generative Adversarial Networks (GANs)} \label{sectionGAN}

Generative Adversarial Networks are one of the most exciting and most promising innovations in machine learning that learn the patterns and statistical distributions of the input data in order to synthesize new samples \cite{goodfellow2014generative} that do not exist in the original data. The synthesizing ability of GANs has been used in semantic image editing \cite{bau2020semantic}, data augmentation \cite{antoniou2017data}, and style transfer \cite{zhu2017unpaired}. Furthermore, one can use the patterns learned by GANs for other data-intensive tasks, such as classification \cite{zhan2017semisupervised}. 

GANs consist of a generator and a discriminator, which are both neural networks that work together. During  training, the generator learns to generate plausible data, and the discriminator learns to distinguish the generator's fake data from real data and penalizes the generator for producing implausible results. As training progresses, the generator becomes better at generating plausible data, and the discriminator has a harder time distinguishing between real and fake data.  Using GANs in practice must overcome several challenges, including those induced by vanishing gradients, mode collapse, and convergence failures.  A good and recent review on algorithms, theory, and applications of GANs can be found in \cite{gui2020review}.  

\subsubsection*{Transferring GANs}

Deep Neural Networks achieve promising results on image classification problems as long as large datasets are available for training. For example, Xception \cite{chollet2017xception} is trained on 350M images. However, many practical use cases do not have the required abundance of data needed for training of these networks. This is where  transfer learning can come in.

As the name implies, transfer learning reuses the knowledge learned by a source network to improve the performance of a target network trained on a different but related task. Transfer learning can significantly increase the network performance for situations when the data is scarce \cite{Pan2010survey}.  

Recent results of applying transfer learning in conjunction with generative adversarial networks \cite{shin2016generative,wang2018transferring,noguchi2019image} show that by transferring pre-trained knowledge, one can effectively reduce the amount of training data in the context of image generation. In addition, this knowledge increases the GAN convergence rate, and improves the convergence failures due to the well-known mode collapse \cite{goodfellow2014generative}. 


\subsection{Proposed GAN Architecture}
\begin{figure*}[th!]
	\centering
	\includegraphics[width=\textwidth]{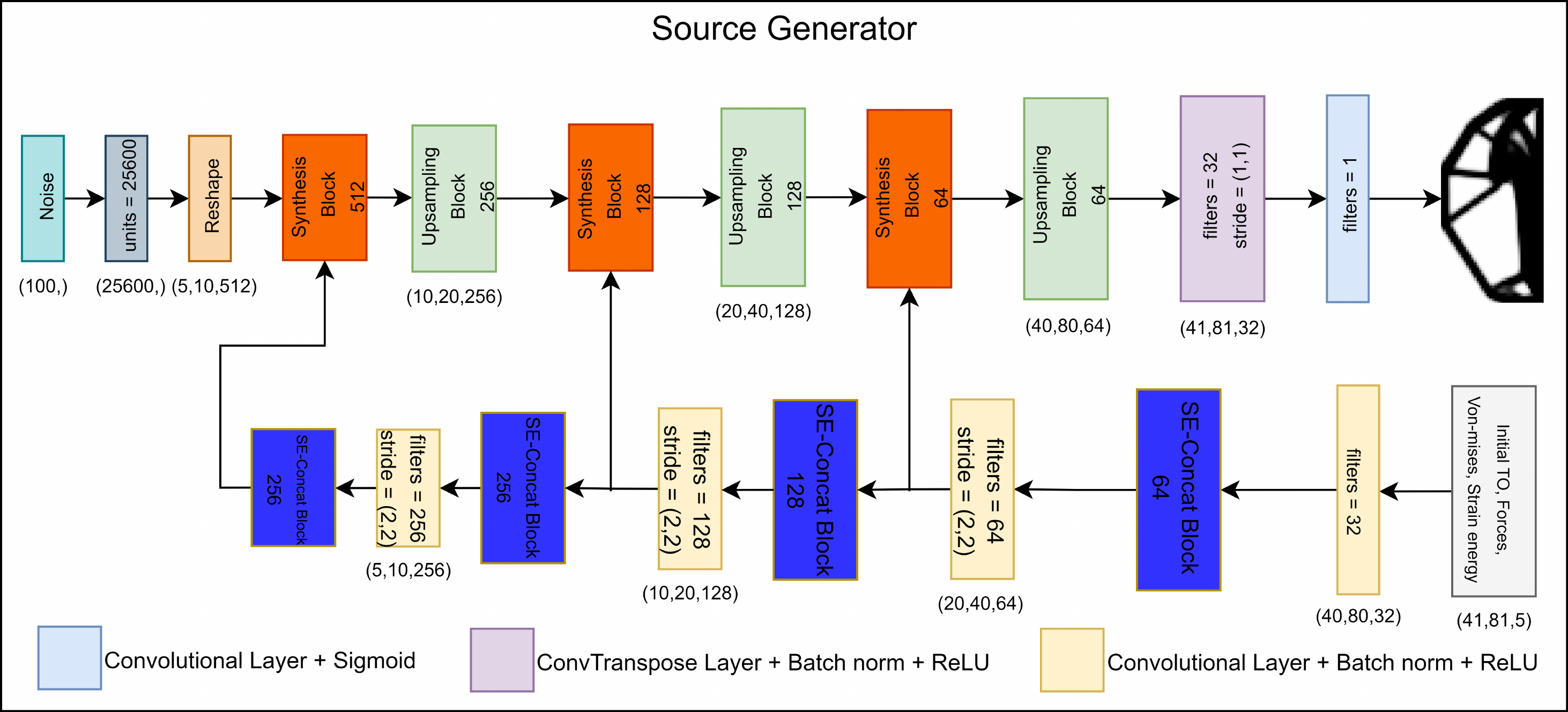}
    \caption{The architecture of the source generator. The numbers below the blocks show the size of the out put of blocks.}
    \label{fig:sourceGEN}
\end{figure*}

\begin{figure*}[th!]
	\centering
	\includegraphics[width=\textwidth]{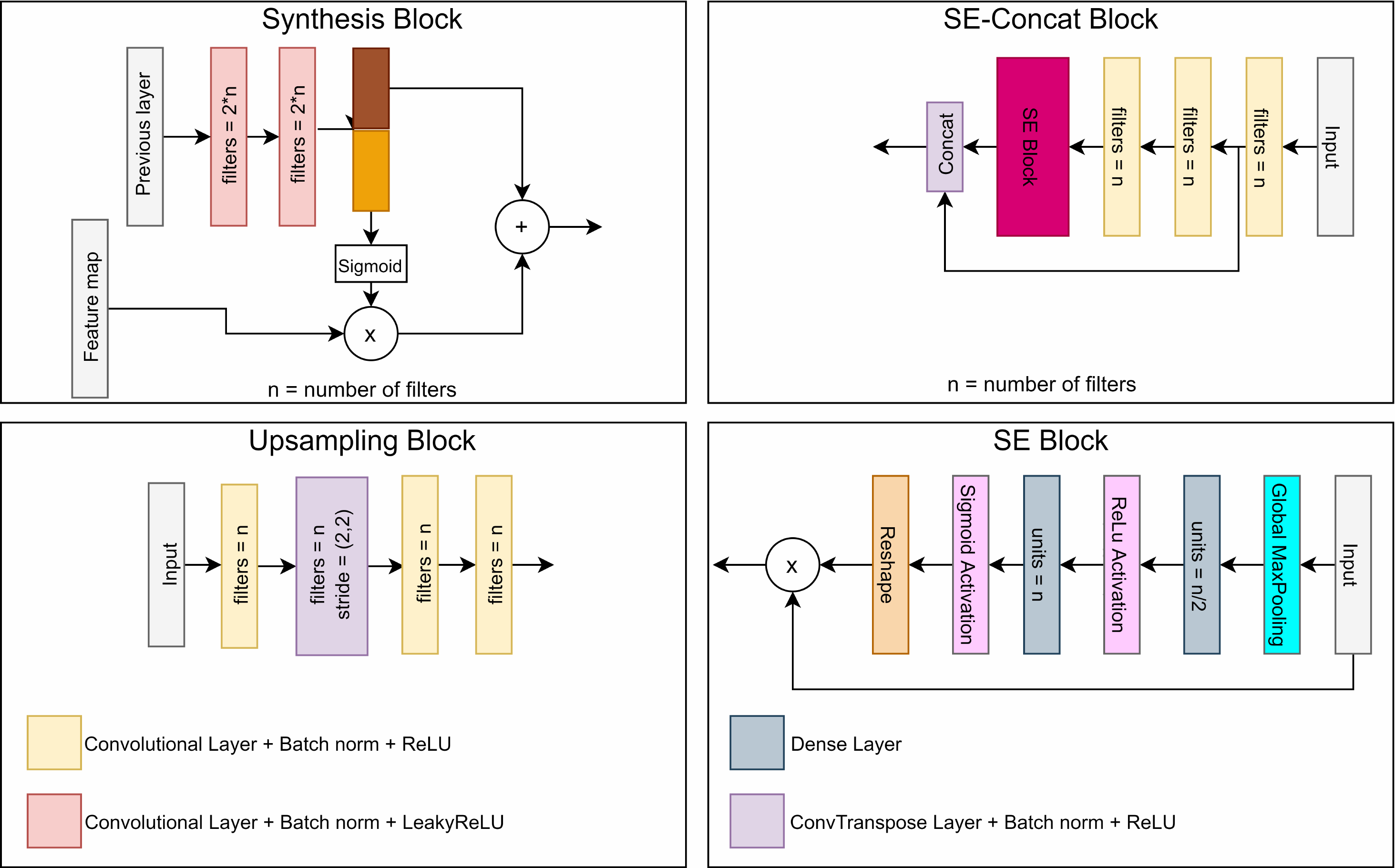}
    \caption{The architecture of the synthesis block, upsampling block, SE-Res block and SE block.}
    \label{fig:blocks}
\end{figure*}

\begin{figure}[th!]
    \centering
    \includegraphics[width=\columnwidth]{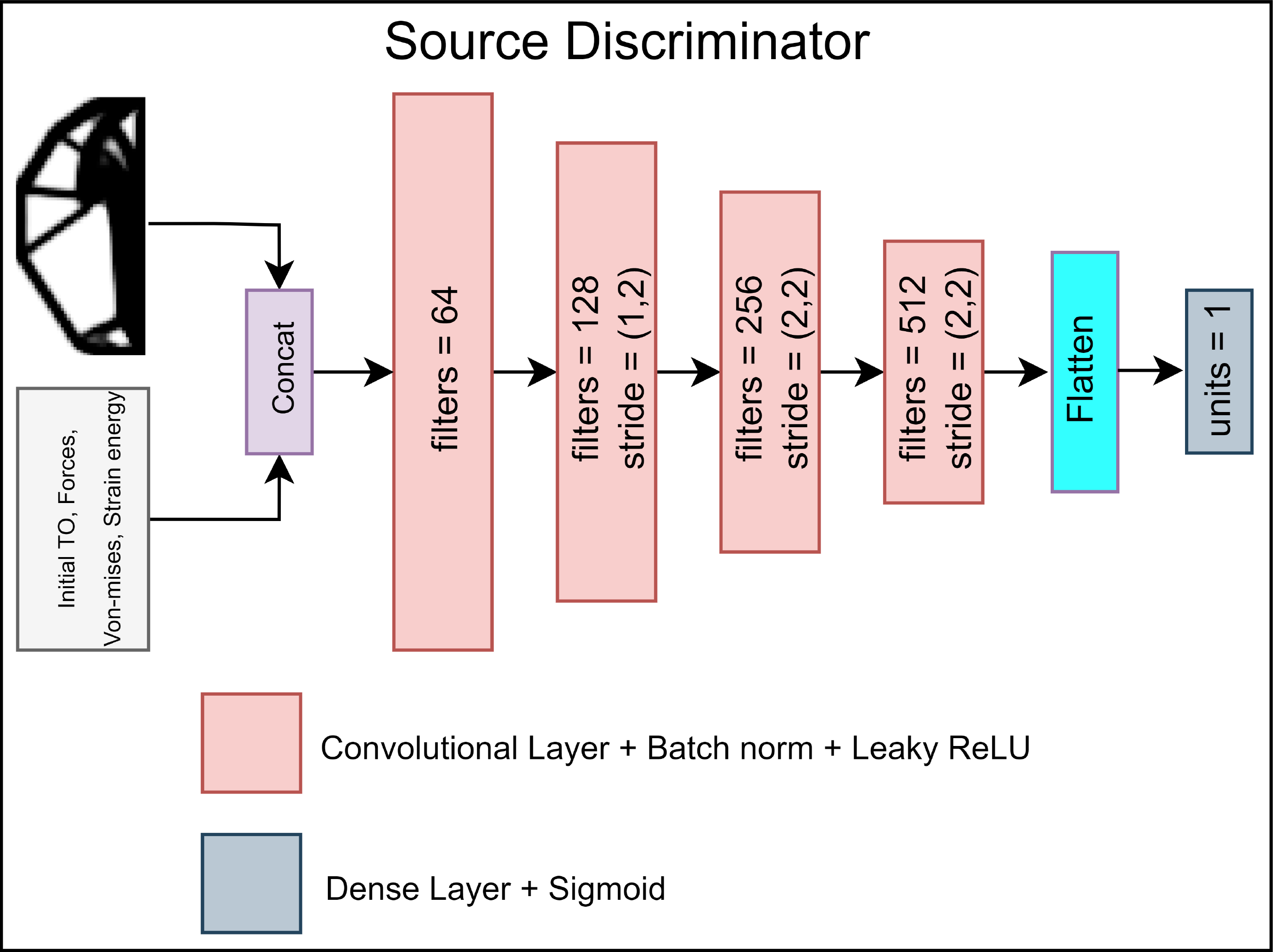}
    \caption{The structure of the source discriminator.}
    \label{fig:sourceDIS}
\end{figure}

\begin{figure}[th!]
    \centering
    \includegraphics[width=\columnwidth]{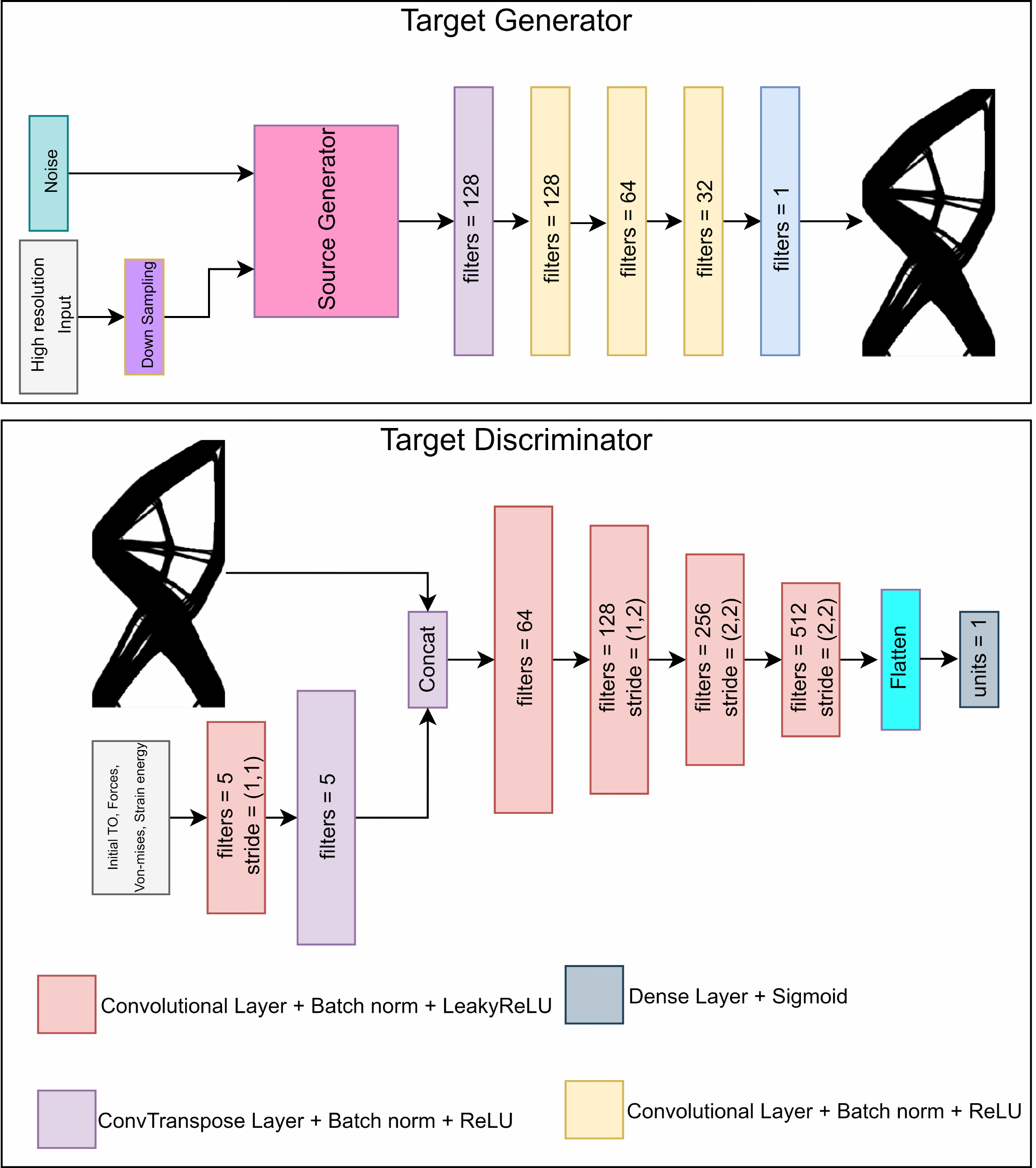}
    \caption{The architecture of the target generator and discriminator.}
    \label{fig:Targetmodel}
\end{figure}

As mentioned above, GANs consist of a generator and a discriminator, and in our case both contain a source as well as a target neural networks. The overall architecture of the source generator is inspired by \cite{park2018mc} and separates the subnets handling the feature extraction from the synthesis. The synthesis subnet  is responsible for generating the optimum structure from a noise vector using the feature maps obtained by the feature extraction subnet. The latter is inspired by SE-Res Net introduced in \cite{hu2018squeeze}, but we remove the addition layer \cite{he2016deep} from the SE-Res block and use a concatenation layer instead to increase the information flow between the layers and reuse the extracted features \cite{huang2017densely}.

The architecture of the source generator is shown in Figure \ref{fig:sourceGEN}, and the corresponding input data is described in the next section. The synthesis subnet uses a series of synthesis blocks described in Fig.\ref{fig:blocks} that helps to combine the feature maps of boundary conditions with the previous layer and determine the amount of the feature information that should be retained for the following blocks.  The SE-Concat block contains a squeeze and excitation (SE) block \cite{hu2018squeeze}, which improves channel interdependencies and determines the important features of the input, as shown in Figure. \ref{fig:blocks}.  

The source discriminator is built using convolutional layers and a dense layer to decide whether an optimum structure output by the generator is fake, as  shown in Fig.\ref{fig:sourceDIS}. The discriminator's inputs are a predicted structure as well as its corresponding boundary information, and the output is $0$ for the fake structures and $1$ for the real structures.

The target generator is built by adding a convolutional transpose layer and four convolutional layers on top of the source generator. Before using the source generator on the target GAN, we remove the last two-layers of the source generator. Furthermore, we add a downsampling function at the front of the target generator to reduce the high resolution input dimension to the low-resolution input required by the pre-trained generator. The architecture of the target discriminator is very similar with that of the source discriminator, with the exception of an additional convolutional layer and of a convolutional transpose layer tasked with matching the output resolution. The structure of the target generator and discriminator are shown in Fig.\ref{fig:Targetmodel}. 

\subsection{Dataset}

We used a SIMP-based topology optimization code \cite{andreassen2011efficient} to generate our training and test cases. The code is modified to automate the data generation for different domains and boundary conditions. The inputs are the voxelized domain geometry,  volume fraction, filter radius, load, and displacement boundary conditions.  The volume fraction and filter radius are prescribed to 0.5 and 1.5, respectively.

\begin{figure*}[ht!]
	\centering
	\includegraphics[width=\textwidth]{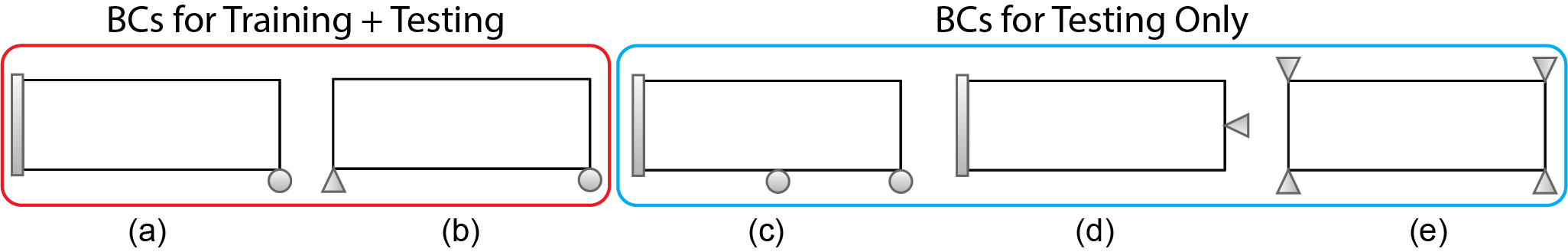}
    \caption{Boundary conditions: (a) and (b) are used for training and testing; (c) -- (e) are unseen BCs only used for testing.}
    \label{fig:BCBeam}
\end{figure*}

The magnitudes of the force components were sampled using uniform random sampling within the range $[-100,100]$N. The location of the external load is selected according to a uniform random sampling within prescribed ranges along the coordinate axes. For example, the force components $F_x$ and $F_y$ for a 2D beam domain are applied within the range $[\frac{b_x}{2},b_x]$ and $[0, b_y]$, respectively, where $b_x$ and $b_y$ are the beam dimensions in the $x$ and $y$ directions. We also used a discrete random sampling to select one of the displacement boundary conditions shown in fig. \ref{fig:BCBeam} (a) and (b).

Our inputs were captured in five channels for the 2D cases represented as matrices:
\begin{enumerate}
    \item First channel: Initial density value for each voxel.
    \item Second channel: Force component in the $x$ direction at each voxel,
    \item Third channel: Force component in the $y$ direction at each voxel,
    \item Fourth channel: The von Mises stress of the initial domain defined as:\\ \\
             $\sigma_v = \sqrt{\sigma_x^2 + \sigma_y^2 - \sigma_x \sigma_y + 3\sigma_{xy}^2}$         
    \item Fifth channel: The strain energy density function of the initial domain defined as:\\ \\
    $ W =\displaystyle \frac{\sigma_x \epsilon_x + \sigma_y \epsilon_y + 2\sigma_{xy} \epsilon_{xy}}{2} $\\ \\
\end{enumerate}
where $\sigma_i$ and $\epsilon_i$ are stress and strain in direction $i$, respectively. The stress and strain corresponding to the initial domain were computed with SolidsPy library \cite{solidspy} in Python.
    
Based on the information given above, a low resolution dataset (11,000 cases) and a much smaller high resolution dataset (1500 data) were generated for the source and target GAN.

\subsection{Training Process and Performance Metrics}\label{training:sec}
The generator and discriminator of the GAN should be trained simultaneously, and we followed the following training process: (1) trained the discriminator for one batch, (2) trained the generator for one batch, (3) repeat step (1) and (2) to finish the batches and go to the next epoch. In the first step, the algorithm generates the labels for the real structures and the fake structures, and then it trains the discriminator on the real structures with label one and the fake structures generated by the generator with label 0. In the next step, the generator is trained on the ground truth data to generate accurate pictures. As the generator getting better at generating the pictures, the discriminator is getting worse in distinguishing between the real and the fake data, and when the accuracy of the discriminator for classifying the data is less than 50\%, the training process finishes.

Our GAN is trained on the domain with boundary conditions shown in fig. \ref{fig:BCBeam}(a) and (b). The loss function for both generator and the discriminator is binary cross-entropy, and Adam is used as the optimizer. We employed the same three criteria that we used in \cite{behzadi2021real} to evaluate the quality of the GAN predictions:
\begin{enumerate}
    \item \textit{Compliance Error}, which is defined as:
    \[Comp. Error =  \frac{|\sum_{j=1}^{M}\sum_{i=1}^{N} (c_{pred}^{ij} - c_{true}^{ij})|}{\sum_{j=1}^{M}\sum_{i=1}^{N} c_{true}^{ij}}       \]  
    where $N$ and $M$ are the number of rows and columns, $c_{pred}^{ij}$ and $c_{true}^{ij}$ are the predicted and reference values of compliance located in the $i^{th}$ row and $j^{th}$ column, respectively. 
    \item \textit{Mean Squared Error} (MSE), which is defined as: 
    \[ MSE = \frac{\sum_{j=1}^{M}\sum_{i=1}^{N} (y_{pred}^{ij} - y_{true}^{ij})^2}{N \cdot M} \]    
    where $y_{pred}^{ij}$ and $y_{true}^{ij}$ are the predicted and reference values of element located in the $i^{th}$ row and $j^{th}$ column, respectively. 
    \item \textit{Binary Accuracy} (BA), which is defined as:
    \[BA = \frac{TP + TN}{N}\]
    where $TP$ (True Positive) is the number of elements correctly predicted as 1, $TN$ (True Negative) is the number of elements  correctly predicted as 0, and $N$ is again the total number of elements. Calculating BA is preceded by rounding the element values to the closest integer (0 or 1). 
    
\end{enumerate}

Additional details about these metrics are provided in  \cite{behzadi2021real}.  All experiments were performed on the UConn HPC facility running Red Hat RHEL7 operating system.

\section{Experiments and Evaluation}\label{results:sec}

\subsection{Evaluation with Traditional Cross-Entropy Loss}

We used a freely available Matlab Code \cite{andreassen2011efficient} to generate 11,000 low-resolution cases (40 x 80) used to train the generator and discriminator of the source GAN and 1500 high-resolution cases to train the generator and discriminator of the target GAN. We augmented the training data by mirroring each SIMP-generated case with respect to the $x$ and $y$ axes. 

\textbf{Training:} The source GAN is trained only once, and the trained source network was reused for various target models. For training our source and target GANs, we used two randomly selected boundary conditions corresponding to those shown in Figure \ref{fig:BCBeam} (a) \& (b). All cases used for training had a single externally applied force with randomized location, orientation, and magnitude within the prescribed ranges. 

\textbf{Testing:} Given the training mentioned above, the remaining 3 boundary conditions illustrated in Figures \ref{fig:BCBeam} (c) -- (e) are therefore unseen to both the source and target GANs and were used exclusively for testing. Also, observe that training was completed with only one external force applied, while we tested cases with 1, 2, 5, and 10 external forces. We extensively tested the performance of our predictions and the generalization abilities of our GANTL framework in the following testing scenarios:

\begin{table}[ht!]
    \centering
    \caption{Testing scenarios for both seen BCs (Fig. \ref{fig:BCBeam} (a), (b)) and unseen (Fig. \ref{fig:BCBeam} (c--e)). Training was performed with one external force, while testing was completed for 1, 2, 5, and 10 randomized external forces.}
    \begin{tabular}{ccccc}
        \toprule
        \footnotesize Scenario \#    &  {\shortstack{\footnotesize BCs from  \\ \footnotesize Figure \#}}  & {\shortstack{\footnotesize \# of external \\ \footnotesize forces}} &  {\shortstack{\footnotesize Predictions \\ \footnotesize in Figure \#}} & {\shortstack{\footnotesize Data \\ \footnotesize in Table \#}} \\
        \midrule
        1 & \ref{fig:BCBeam} (a), (b)  & 1 &  \ref{fig:2Dbeam-rect} &  \ref{tab:2Daccuraciesseen}     \\
        2 & \ref{fig:BCBeam} (c--e)  & 1 &  \ref{fig:2Dbeam-rect-unseen-1force} &  \ref{tab:2Daccuracies-unseen-1force}    \\
        3 & \ref{fig:BCBeam} (a) , (b)   & 2 &  \ref{fig:2Dbeam-rect-seen-2force} &  \ref{tab:2Daccuracies-seen-2force}    \\
        4 & \ref{fig:BCBeam} (c--e)    & 2 &  \ref{fig:2Dbeam-rect-unseen-2force} &  \ref{tab:2Daccuracies-unseen-2force}   \\
        5 & \ref{fig:BCBeam} (a), (b)   & 5 \& 10 &  \ref{fig:5-10 forces} &    -    \\ 
  		\bottomrule
    \end{tabular}
    \label{tab:scenarios}
\end{table} 

\subsection{Discussion}

\begin{figure*}[p!]
    \begin{subfigure}[b]{.48\textwidth}
    \centering
      \includegraphics[width=.5\linewidth]{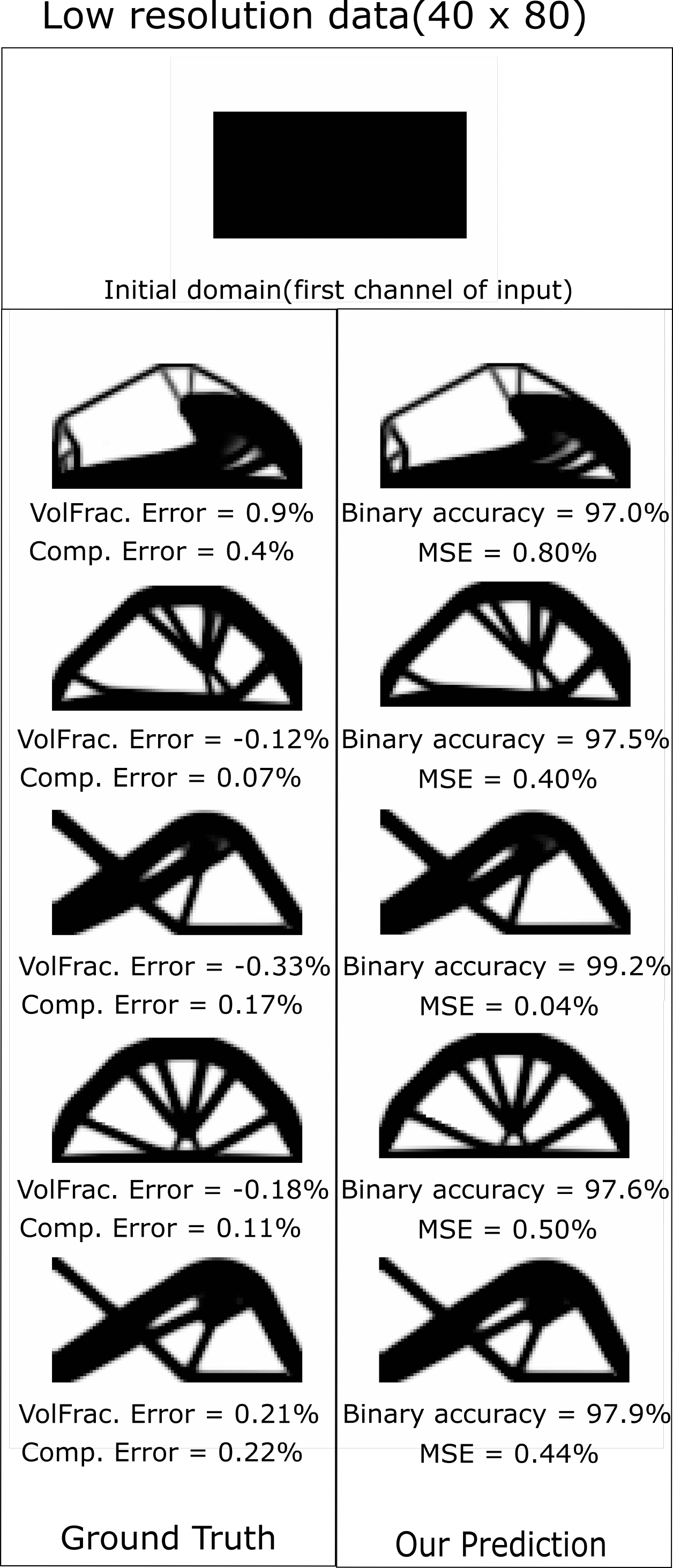}
      \caption{}
      \label{fig:2Dbeam-recta}
    \end{subfigure}
\hspace{-4.3cm}
    \begin{subfigure}[b]{.48\textwidth}
    \centering
      \includegraphics[width=.5\linewidth]{2D/80160.pdf}
      \caption{}
      \label{fig:2Dbeam-rectb}
    \end{subfigure}
\hspace{-4.3cm}
    \begin{subfigure}[b]{.48\textwidth}
    \centering
      \includegraphics[width=.5\linewidth]{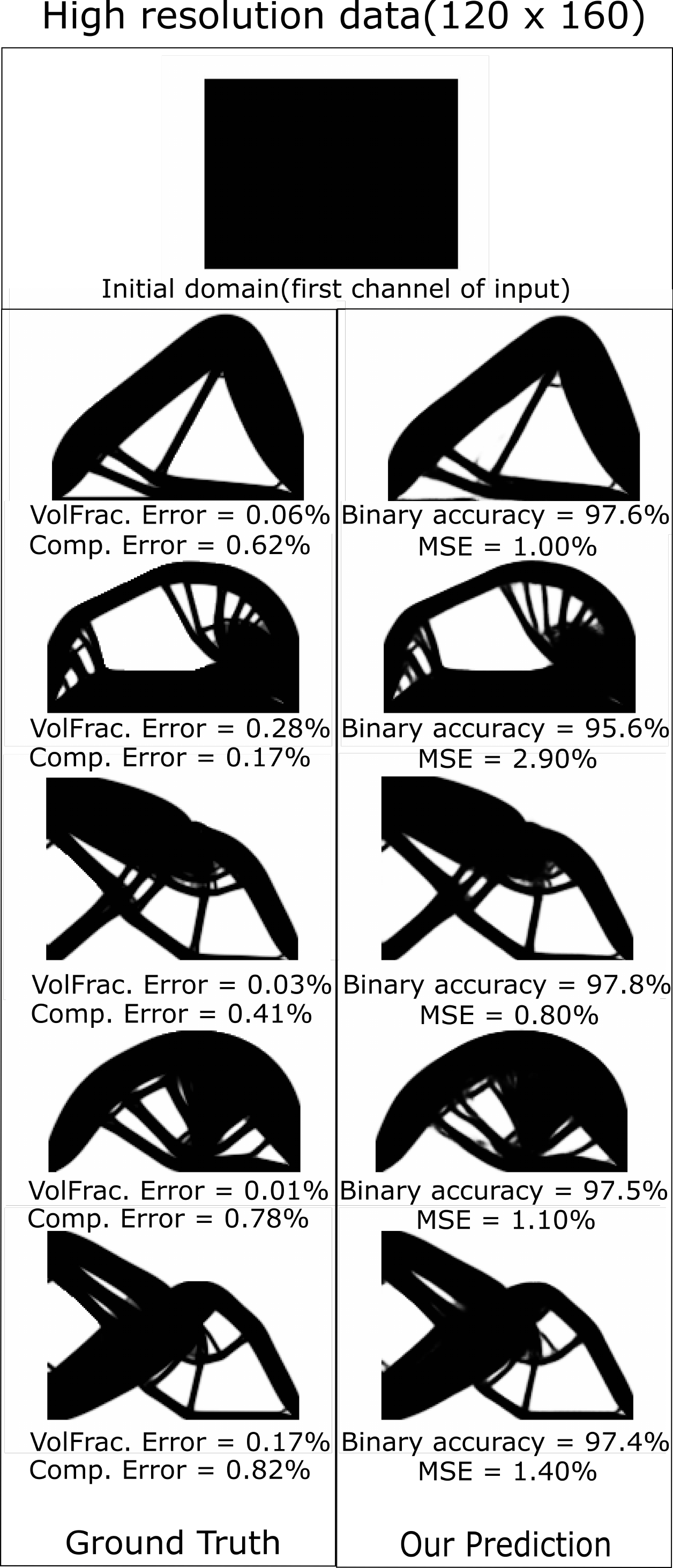}
      \caption{}
      \label{fig:2Dbeam-rectc}
    \end{subfigure}
    \begin{subfigure}[b]{.48\textwidth}
    \centering
      \includegraphics[width=.5\linewidth]{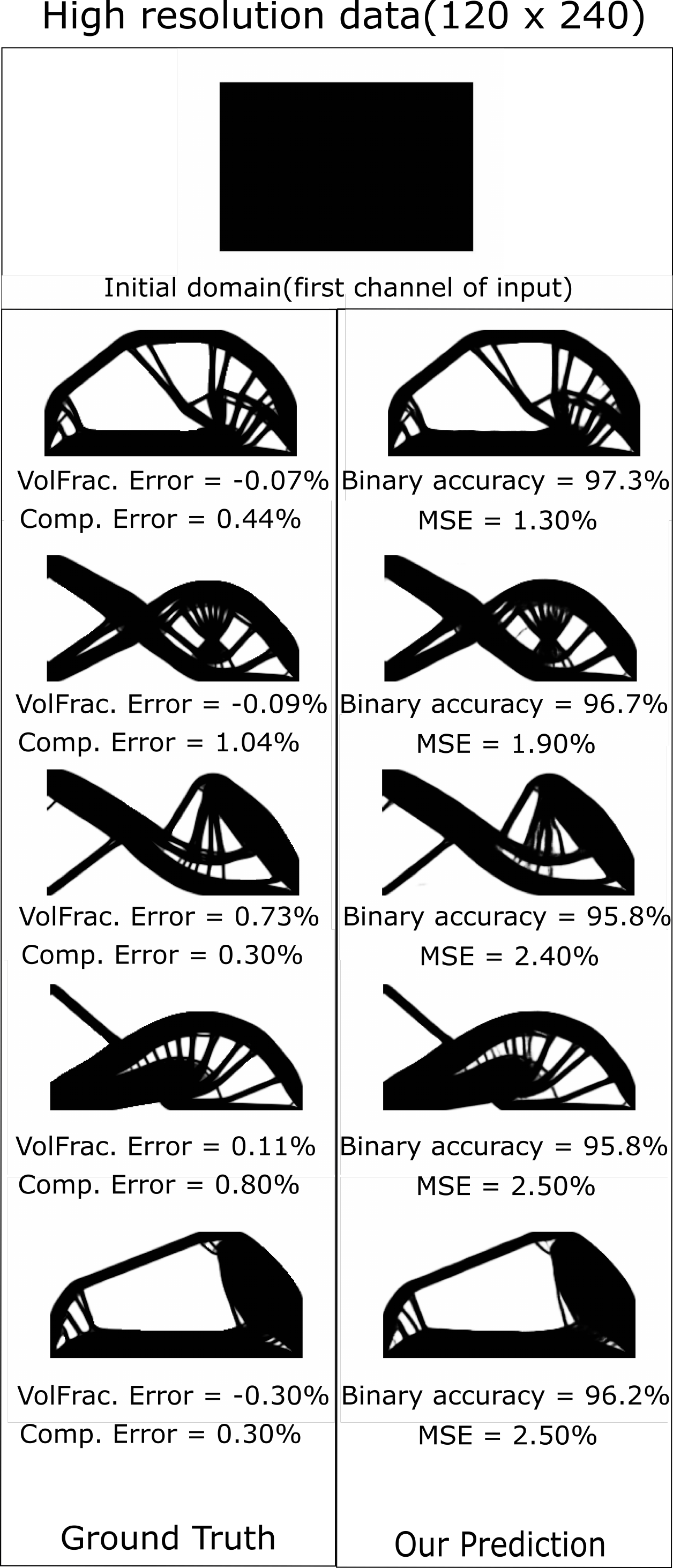}
      \caption{}
      \label{fig:2Dbeam-rectd}
    \end{subfigure}
\hspace{-4.0cm}
    \begin{subfigure}[b]{.48\textwidth}
    \centering
      \includegraphics[width=.5\linewidth]{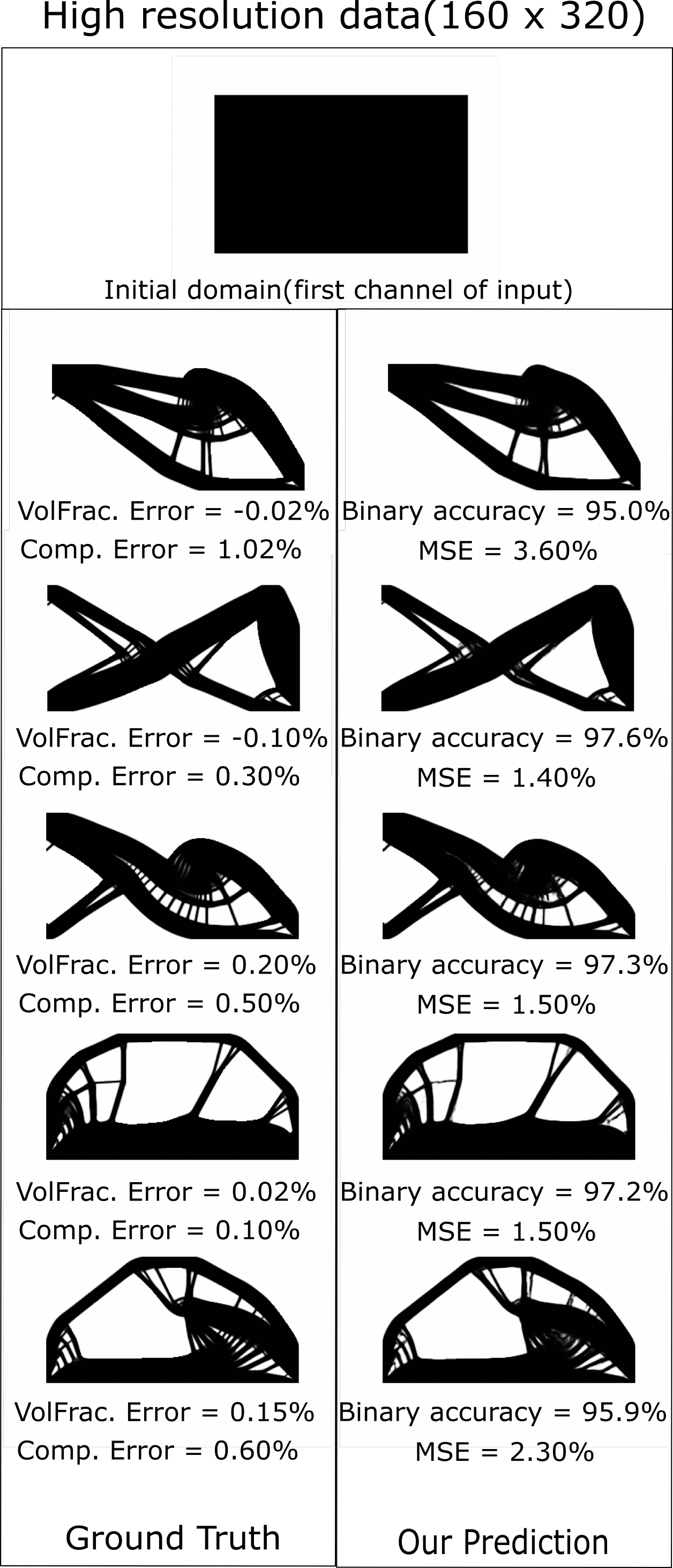}
      \caption{}
      \label{fig:2Dbeam-recte}
    \end{subfigure}
\hspace{-4.0cm}
    \begin{subfigure}[b]{.48\textwidth}
    \centering
      \includegraphics[width=.5\linewidth]{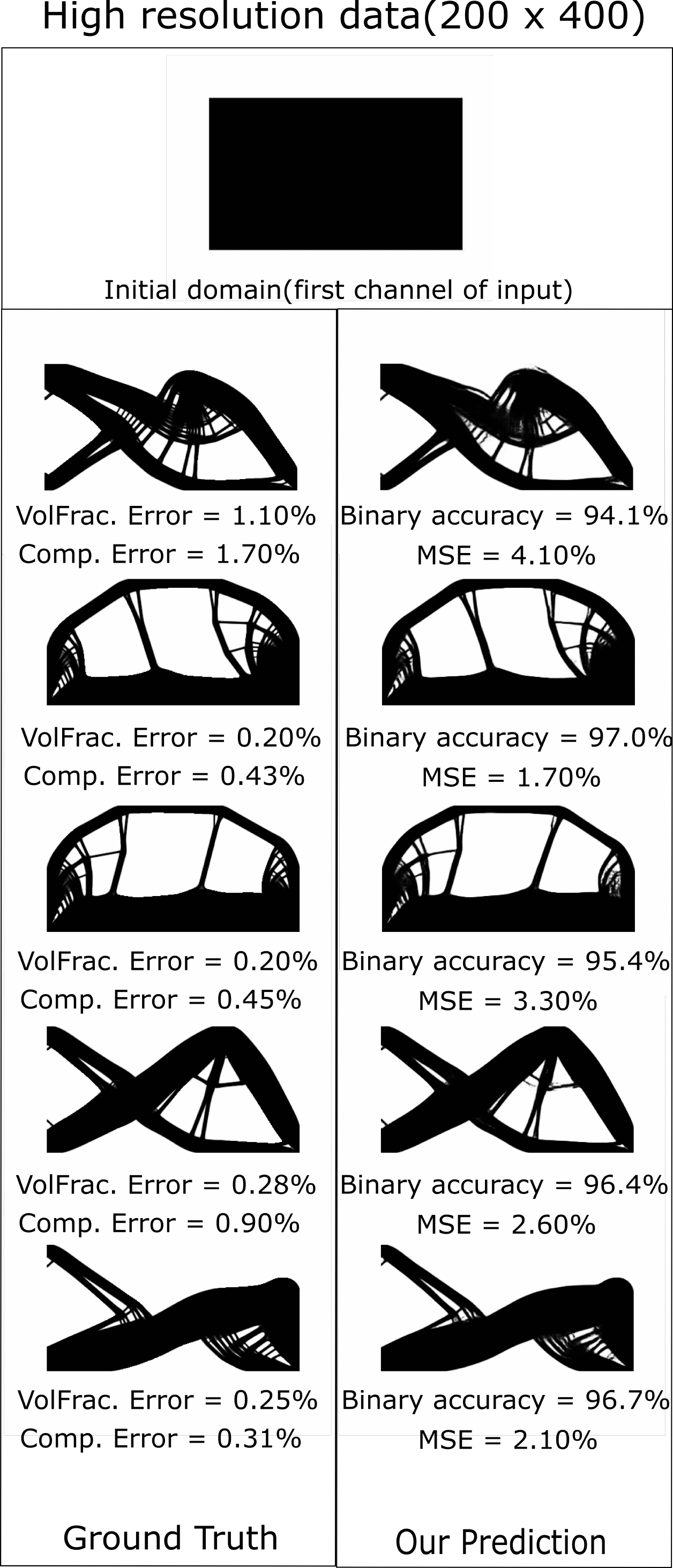}
      \caption{}
      \label{fig:2Dbeam-rectf}
    \end{subfigure}
\caption{Predictions vs. ground Truth for \textit{seen} boundary conditions and one external force:  (a) the prediction of our source GAN alone; (b-f) optimal structures output by the fine tuned target model. The corresponding quality metrics are presented in Table \ref{tab:2Daccuraciesseen}.}
\label{fig:2Dbeam-rect}
\end{figure*}

\begin{figure*}[t!]
    \begin{subfigure}[t]{.48\textwidth}
    \centering
      \includegraphics[width=.5\linewidth]{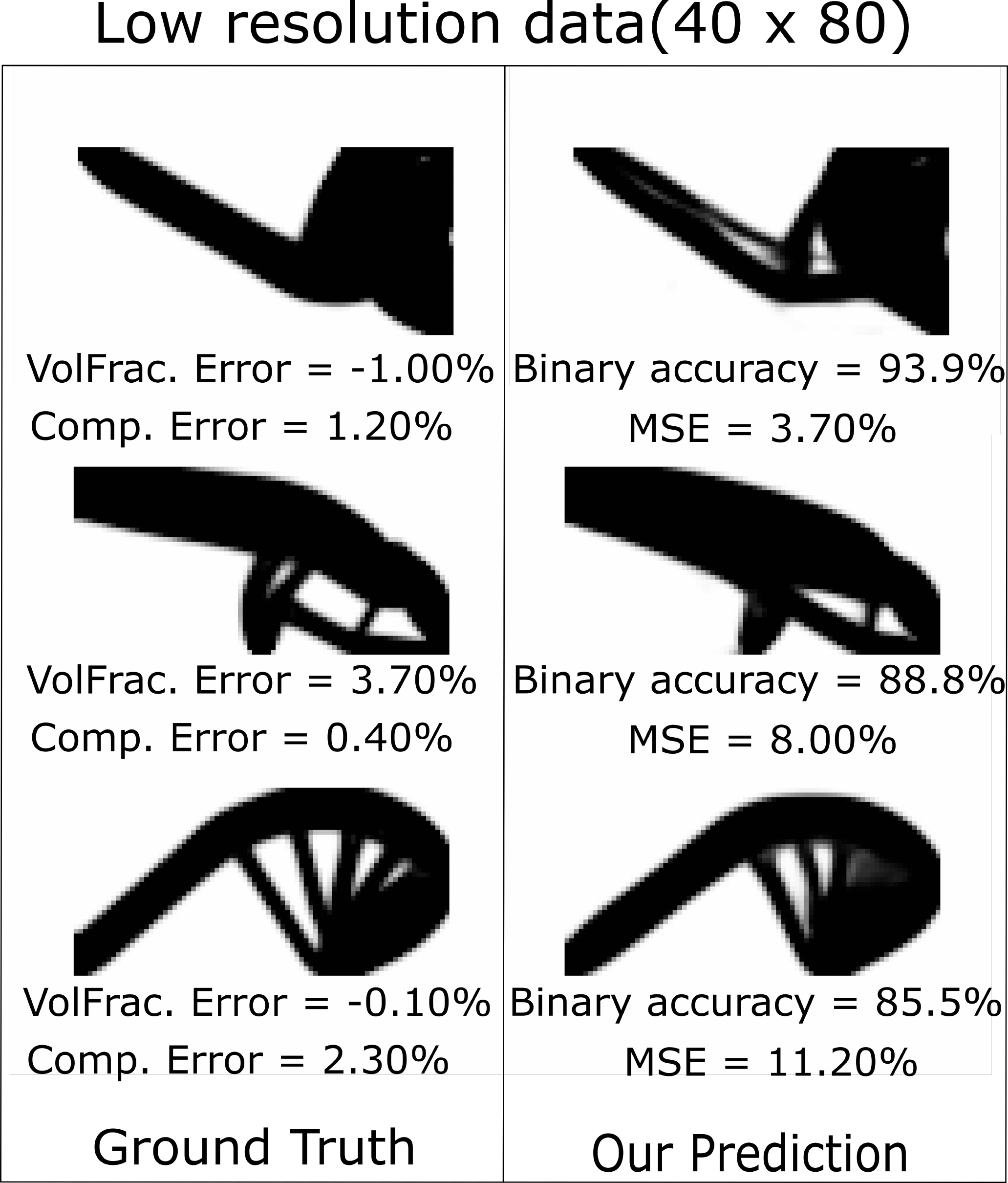}
      \caption{}
      \label{fig:2Dbeam-rect-unseen-1forcea}
    \end{subfigure}
\hspace{-4.3cm}
    \begin{subfigure}[t]{.48\textwidth}
    \centering
      \includegraphics[width=.5\linewidth]{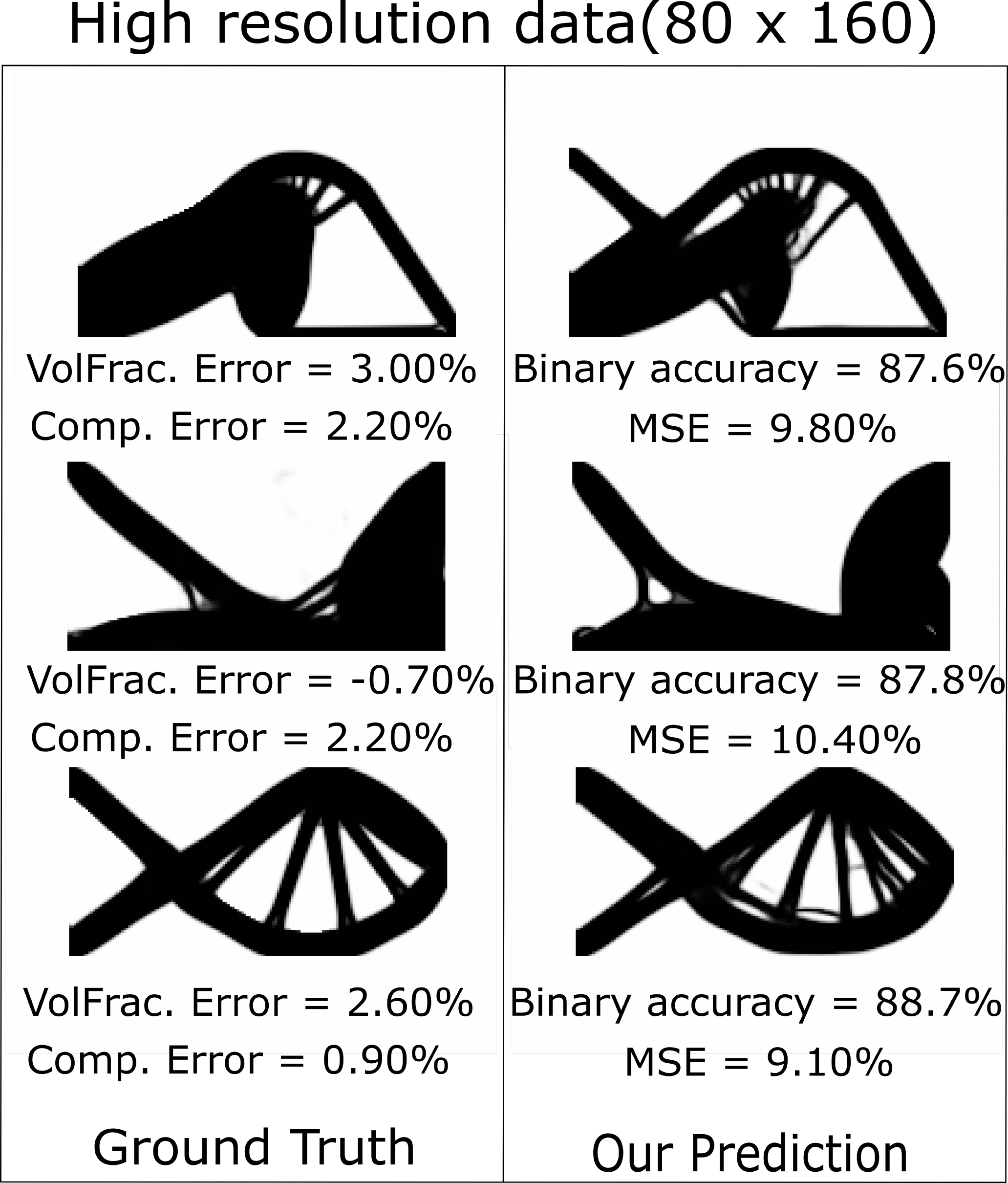}
      \caption{}
      \label{fig:2Dbeam-rect-unseen-1forceb}
    \end{subfigure}
\hspace{-4.3cm}
    \begin{subfigure}[t]{.48\textwidth}
    \centering
      \includegraphics[width=.5\linewidth]{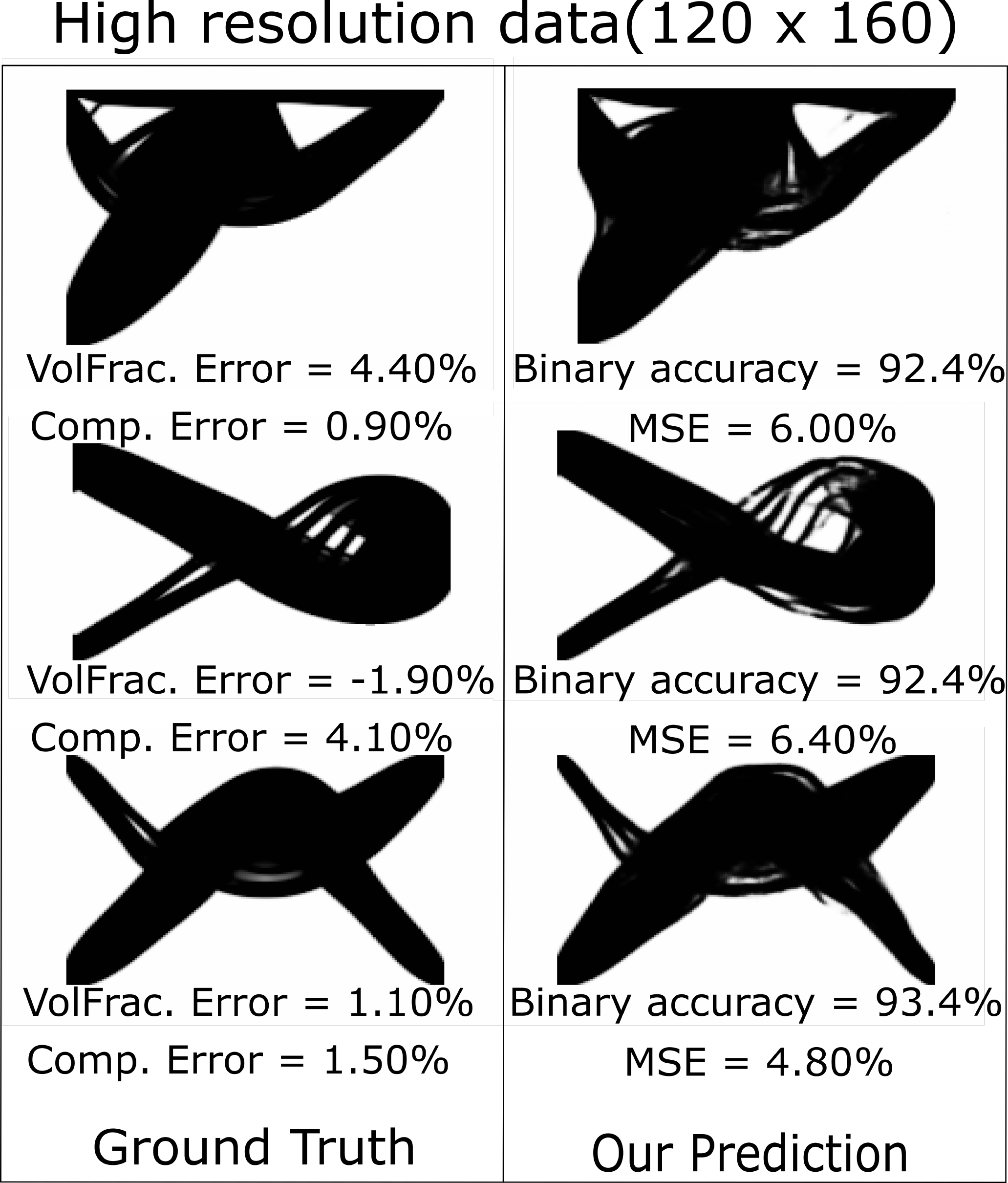}
      \caption{}
      \label{fig:2Dbeam-rect-unseen-1forcec}
    \end{subfigure}
    \begin{subfigure}[t]{.48\textwidth}
    \centering
      \includegraphics[width=.5\linewidth]{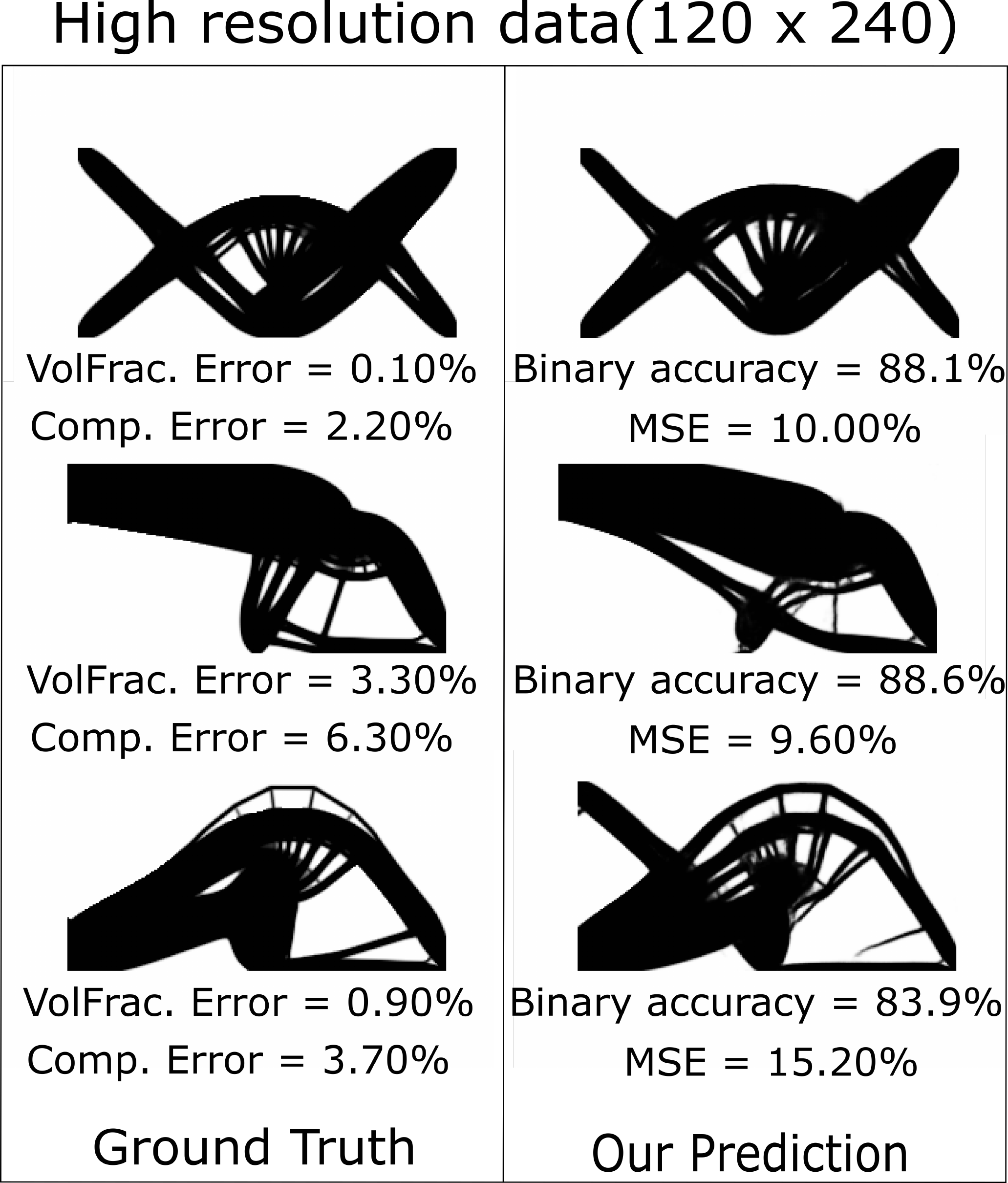}
      \caption{}
      \label{fig:2Dbeam-rect-unseen-1forced}
    \end{subfigure}
\hspace{-4.0cm}
    \begin{subfigure}[t]{.48\textwidth}
    \centering
      \includegraphics[width=.5\linewidth]{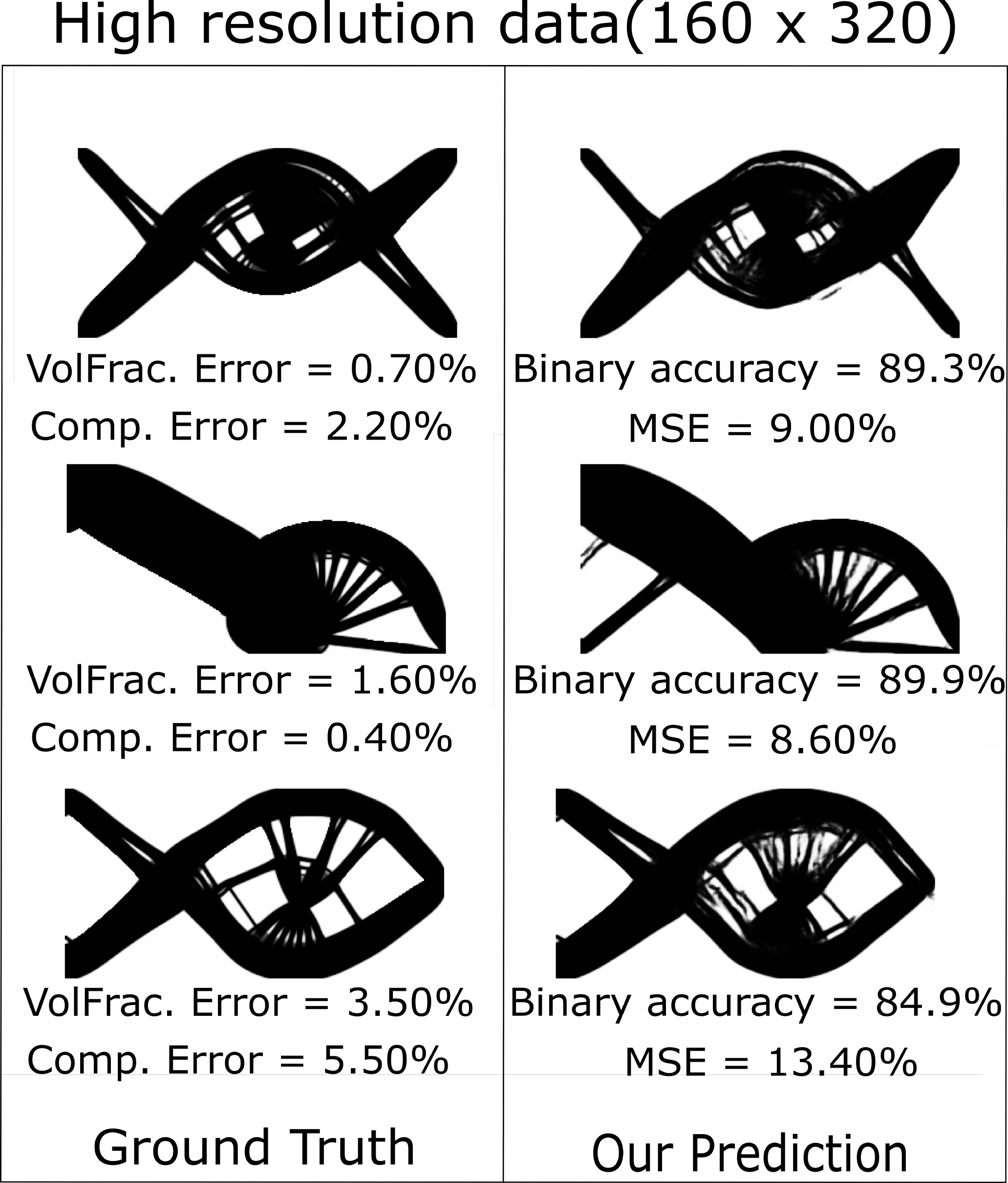}
      \caption{}
      \label{fig:2Dbeam-rect-unseen-1forcee}
    \end{subfigure}
\hspace{-4.0cm}
    \begin{subfigure}[t]{.48\textwidth}
    \centering
      \includegraphics[width=.5\linewidth]{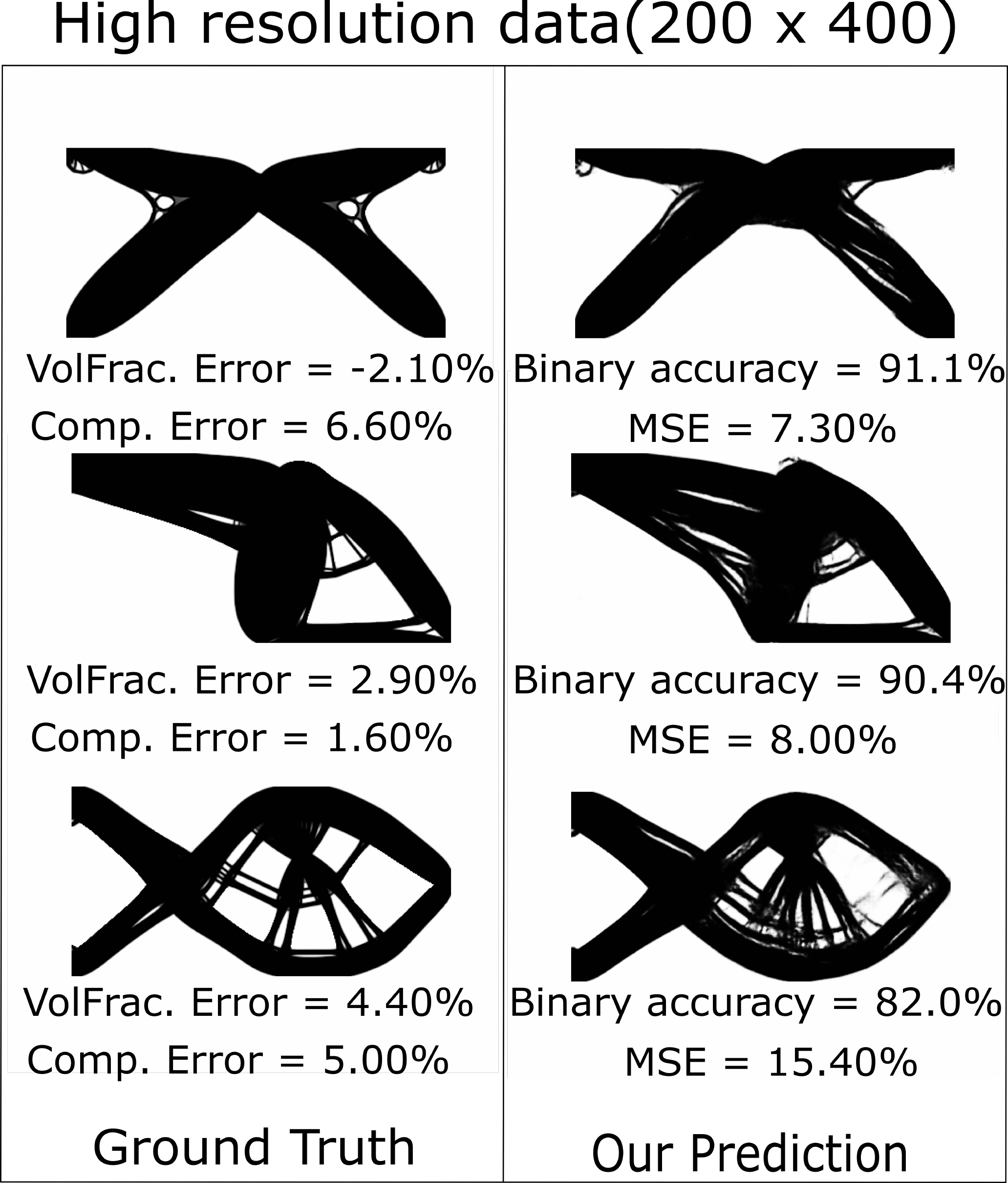}
      \caption{}
      \label{fig:2Dbeam-rect-unseen-1forcef}
    \end{subfigure}
\caption{Predictions vs. ground Truth for \textit{unseen} boundary conditions and one external force:  (a) the prediction of our source GAN alone; (b-f) optimal structures output by the fine tuned target model. The corresponding quality metrics are presented in Table \ref{tab:2Daccuracies-unseen-1force}.}
\label{fig:2Dbeam-rect-unseen-1force}

\end{figure*}

\begin{figure*}[t!]
    \begin{subfigure}[b]{.48\textwidth}
    \centering
      \includegraphics[width=.5\linewidth]{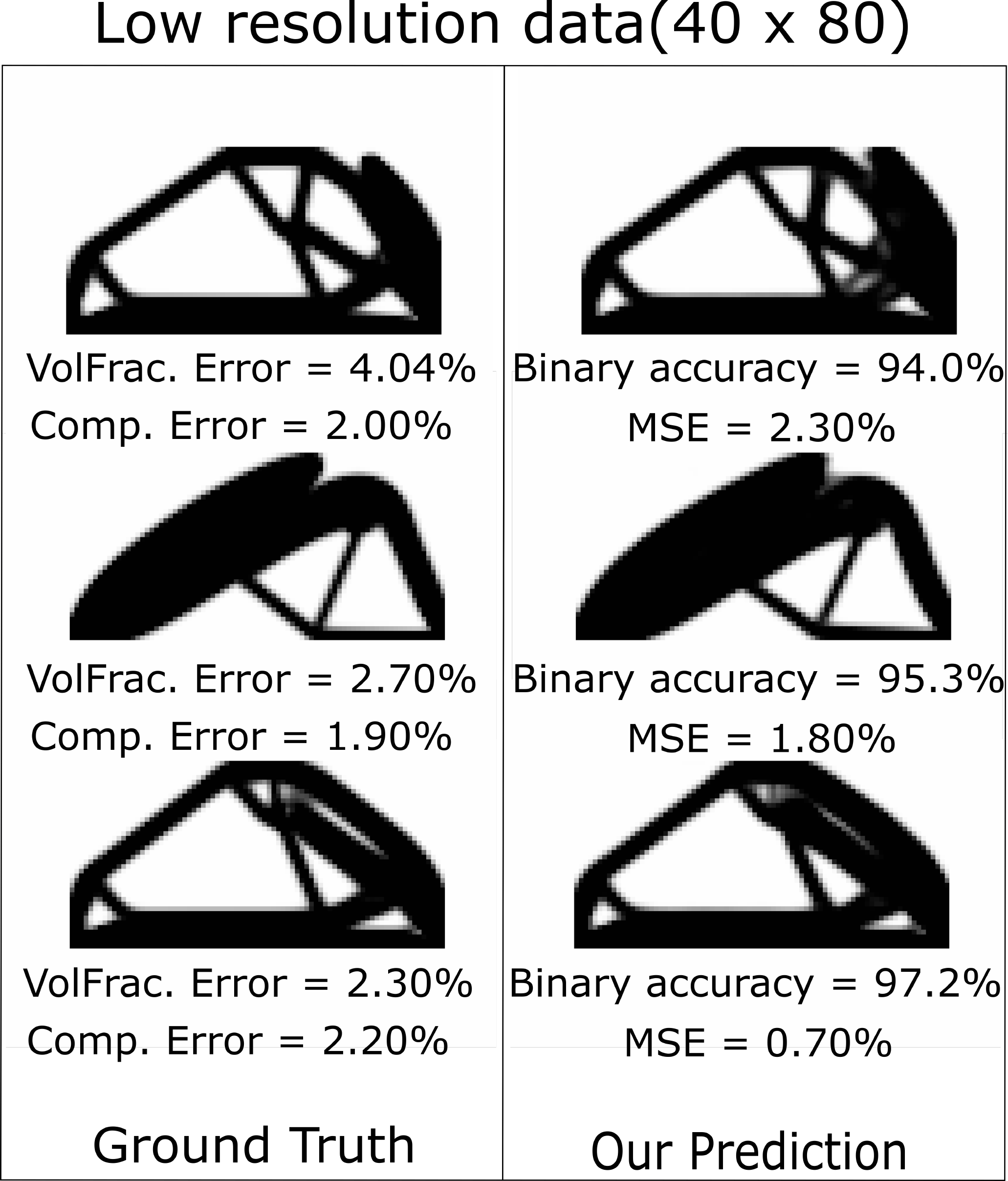}
      \caption{}
      \label{fig:2Dbeam-rect-seen-2forcea}
    \end{subfigure}
\hspace{-4.3cm}
    \begin{subfigure}[b]{.48\textwidth}
    \centering
      \includegraphics[width=.5\linewidth]{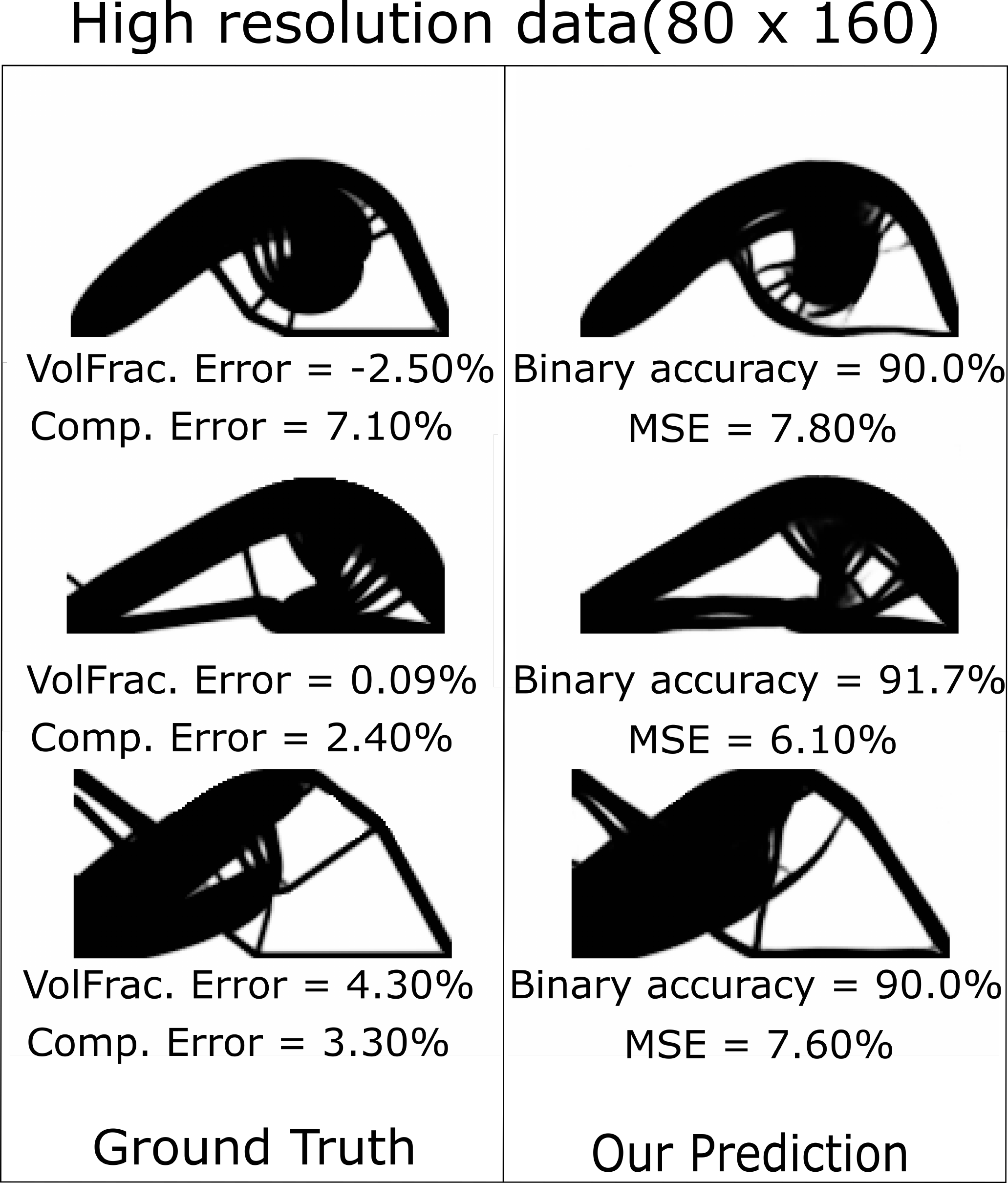}
      \caption{}
      \label{fig:2Dbeam-rect-seen-2forceb}
    \end{subfigure}
\hspace{-4.3cm}
    \begin{subfigure}[b]{.48\textwidth}
    \centering
      \includegraphics[width=.5\linewidth]{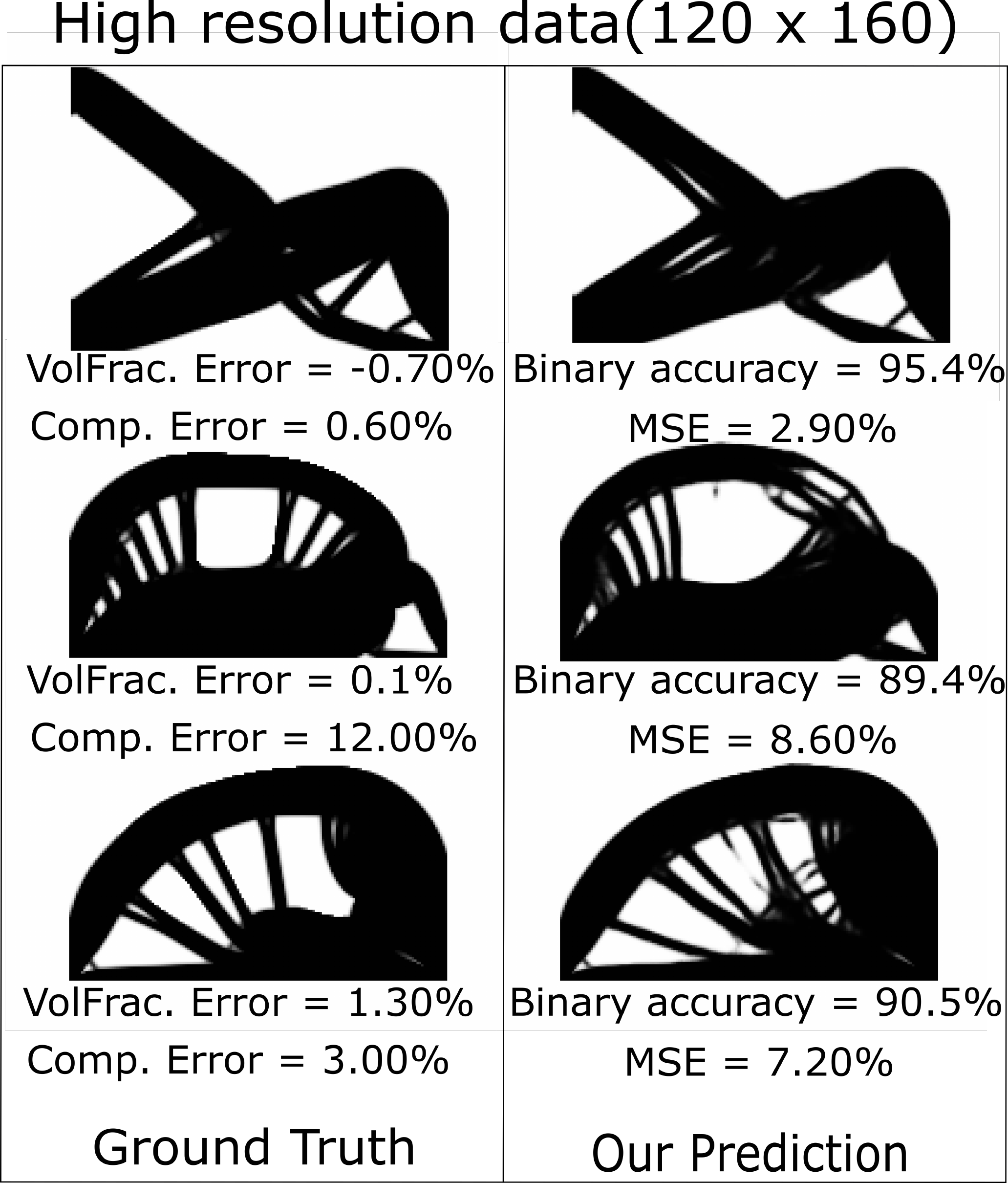}
      \caption{}
      \label{fig:2Dbeam-rect-seen-2forcec}
    \end{subfigure}
    \begin{subfigure}[b]{.48\textwidth}
    \centering
      \includegraphics[width=.5\linewidth]{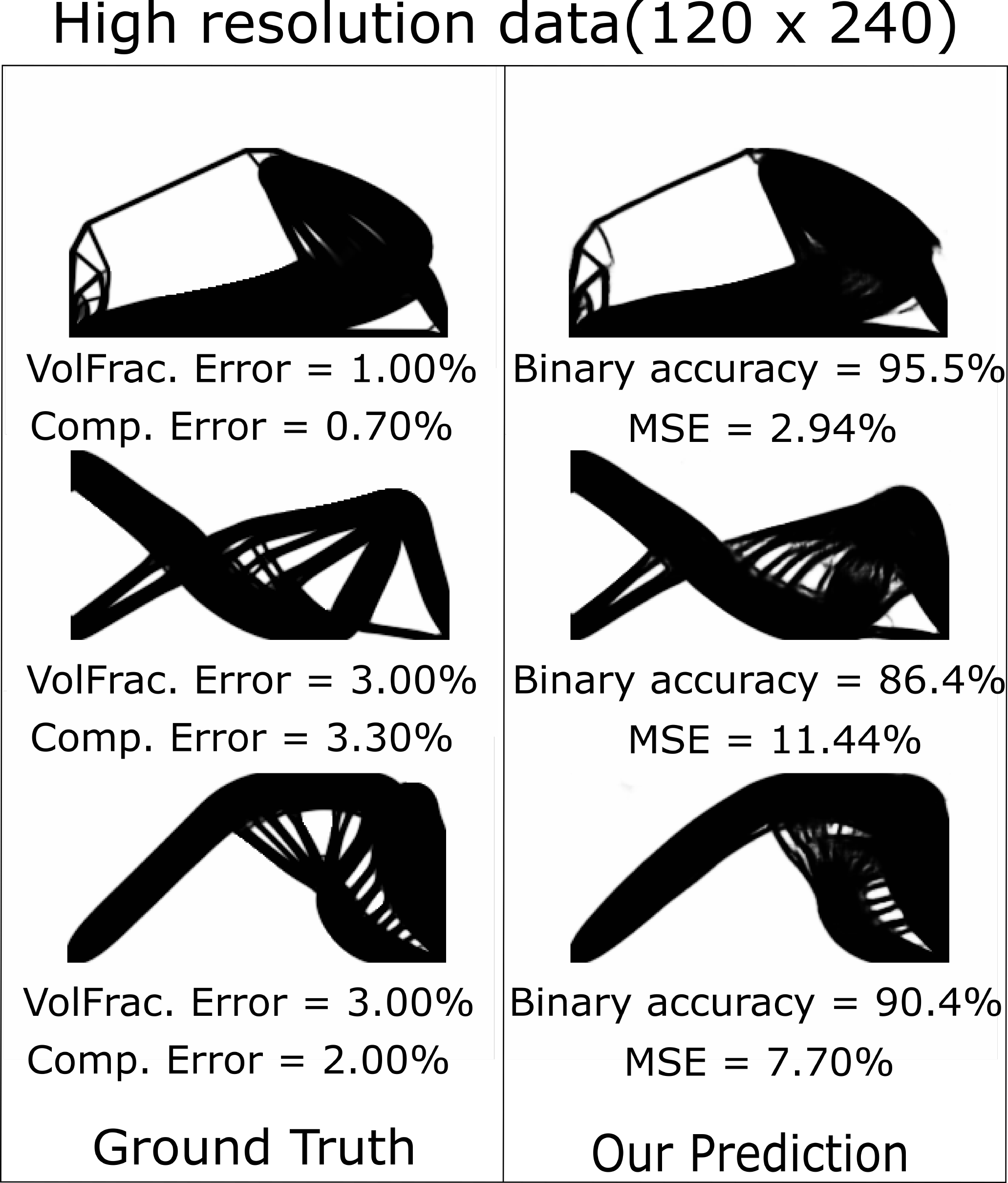}
      \caption{}
      \label{fig:2Dbeam-rect-seen-2forced}
    \end{subfigure}
\hspace{-4.0cm}
    \begin{subfigure}[b]{.48\textwidth}
    \centering
      \includegraphics[width=.5\linewidth]{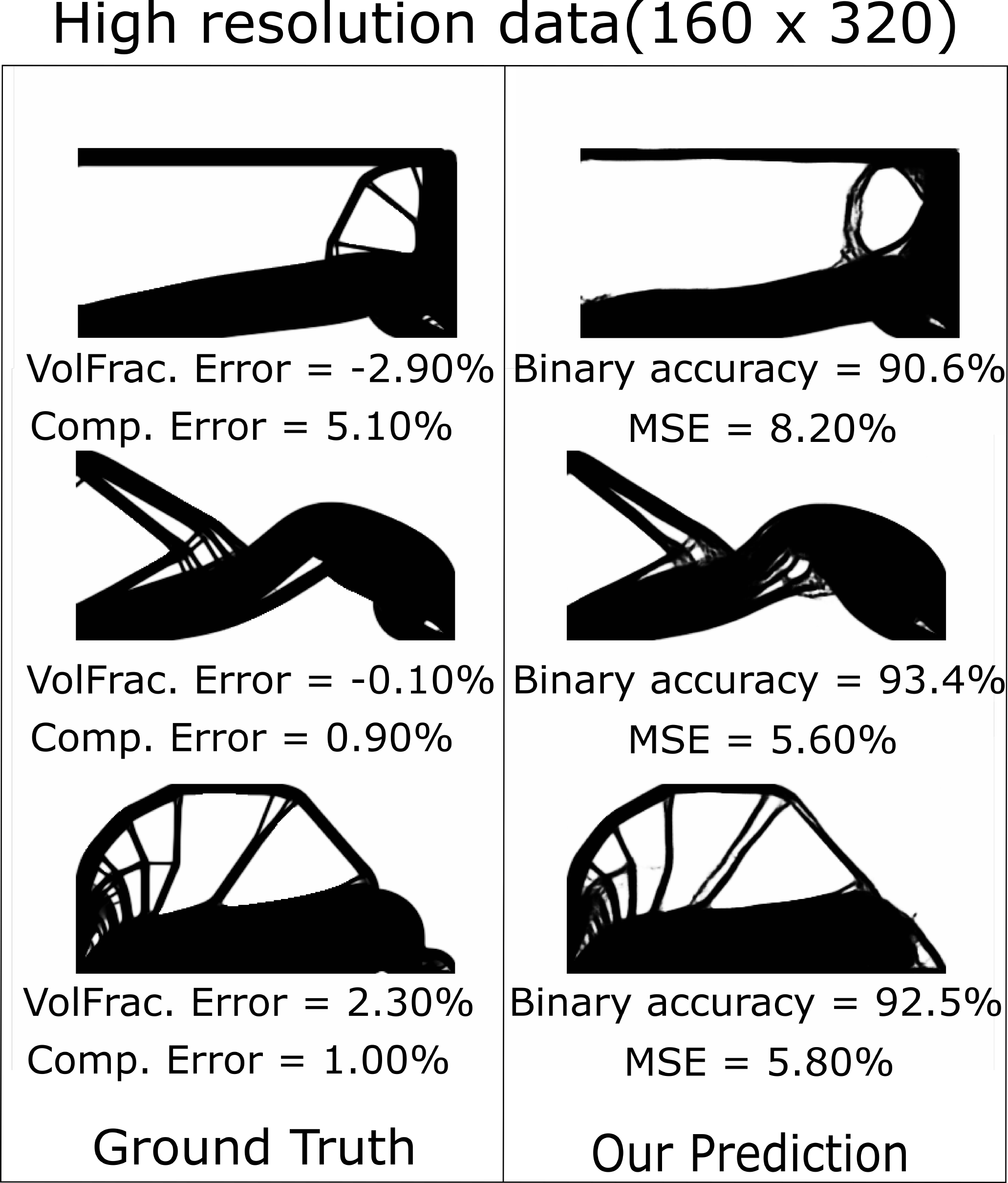}
      \caption{}
      \label{fig:2Dbeam-rect-seen-2forcee}
    \end{subfigure}
\hspace{-4.0cm}
    \begin{subfigure}[b]{.48\textwidth}
    \centering
      \includegraphics[width=.5\linewidth]{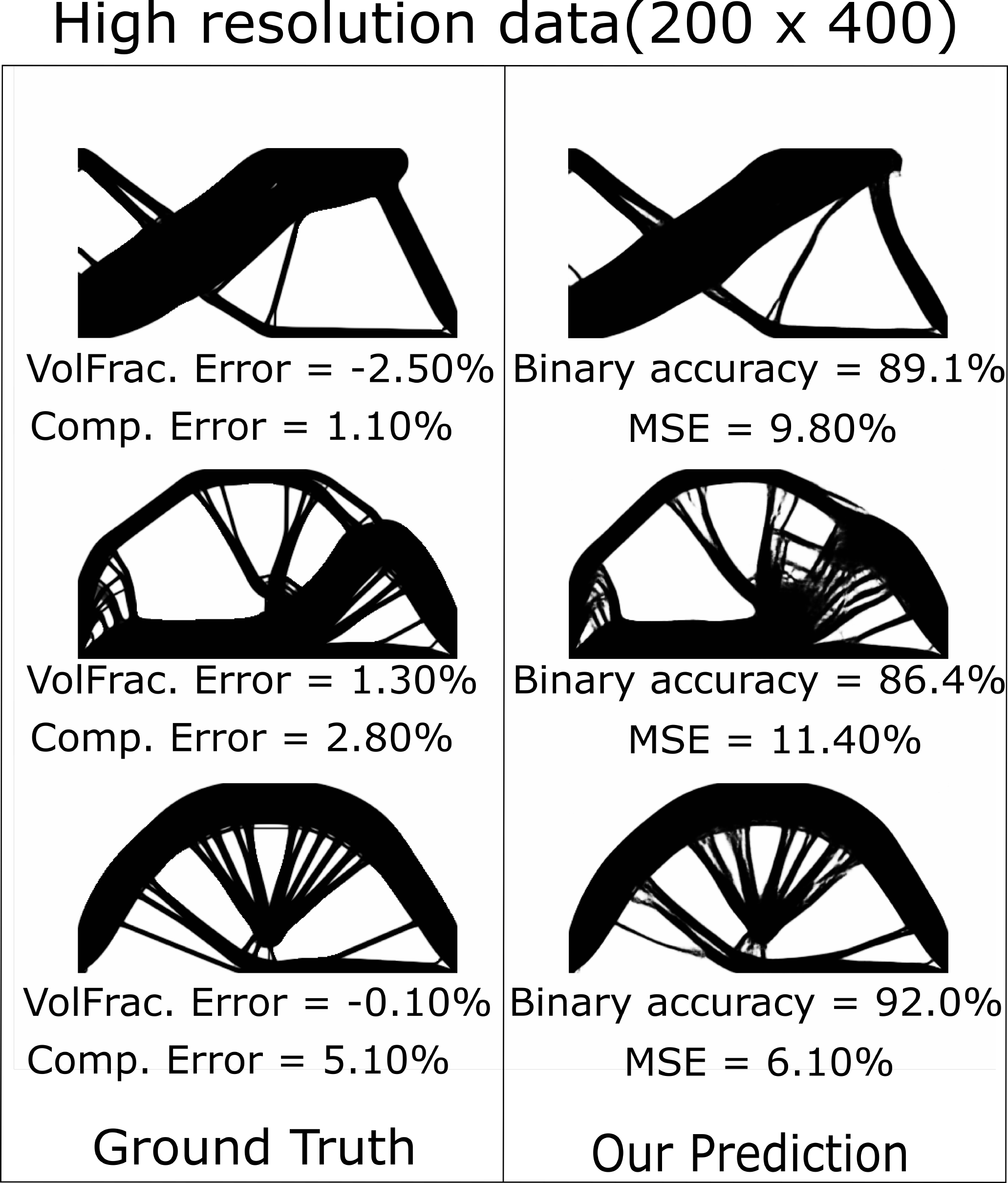}
      \caption{}
      \label{fig:2Dbeam-rect-seen-2forcef}
    \end{subfigure}
\caption{Predictions vs. ground Truth for \textit{seen} boundary conditions and \textit{two external forces}:  (a) the prediction of our source GAN alone; (b-f) optimal structures output by the fine tuned target model. The corresponding quality metrics are presented in Table \ref{tab:2Daccuracies-seen-2force}.}
\label{fig:2Dbeam-rect-seen-2force}

\end{figure*}

\begin{figure*}[t!]
    \begin{subfigure}[b]{.48\textwidth}
    \centering
      \includegraphics[width=.5\linewidth]{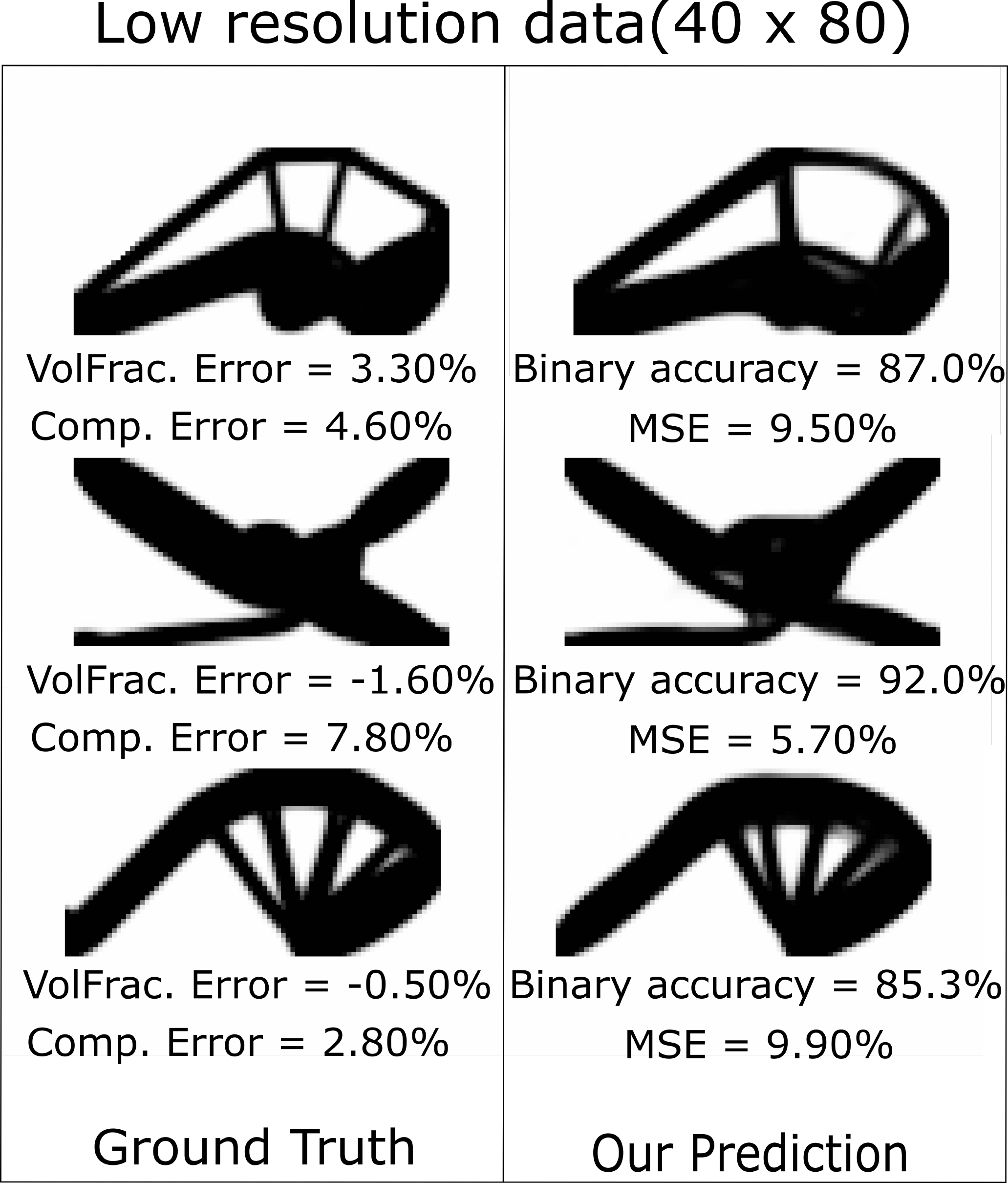}
      \caption{}
      \label{fig:2Dbeam-rect-unseen-2forcea}
    \end{subfigure}
\hspace{-4.3cm}
    \begin{subfigure}[b]{.48\textwidth}
    \centering
      \includegraphics[width=.5\linewidth]{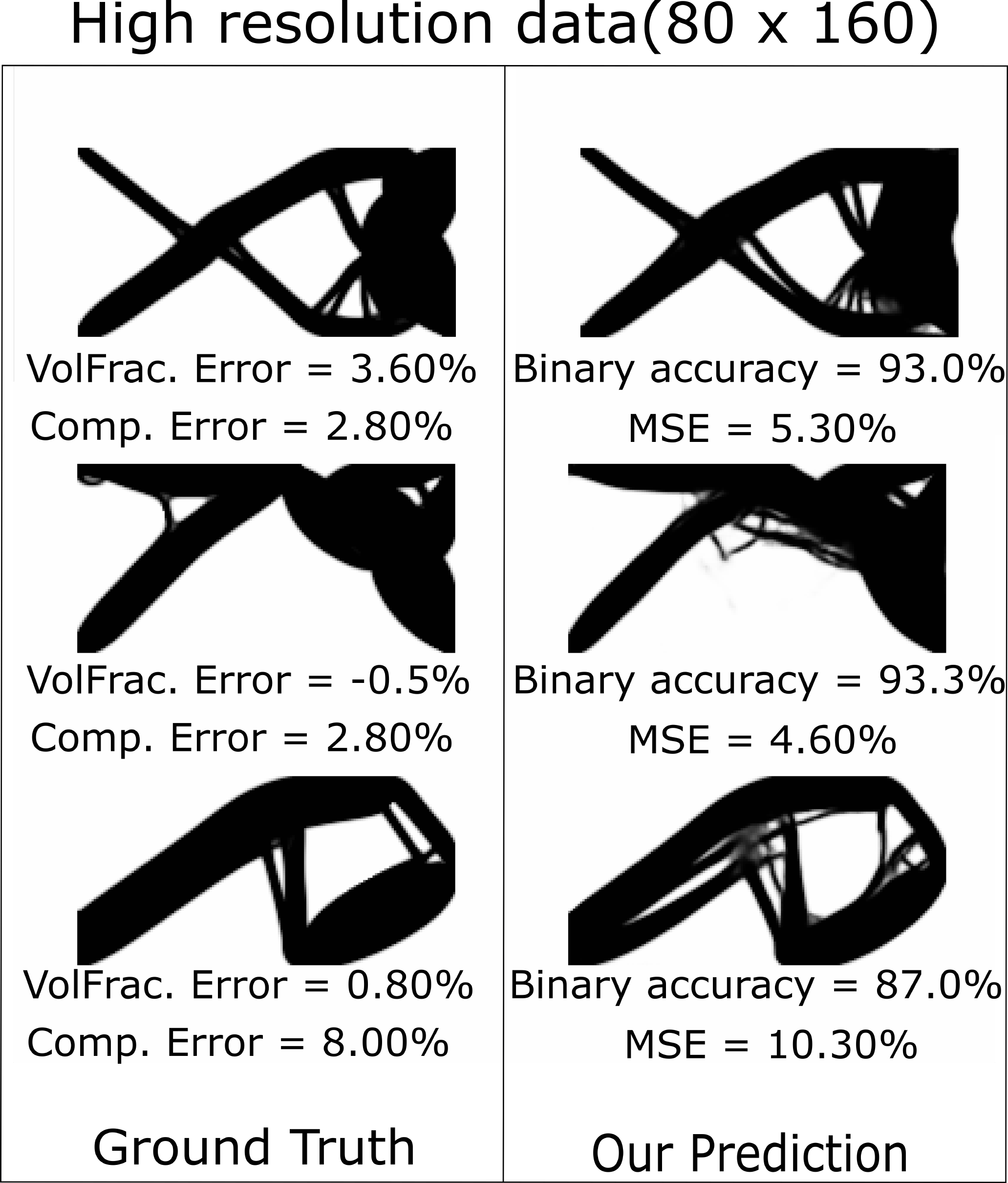}
      \caption{}
      \label{fig:2Dbeam-rect-unseen-2forceb}
    \end{subfigure}
\hspace{-4.3cm}
    \begin{subfigure}[b]{.48\textwidth}
    \centering
      \includegraphics[width=.5\linewidth]{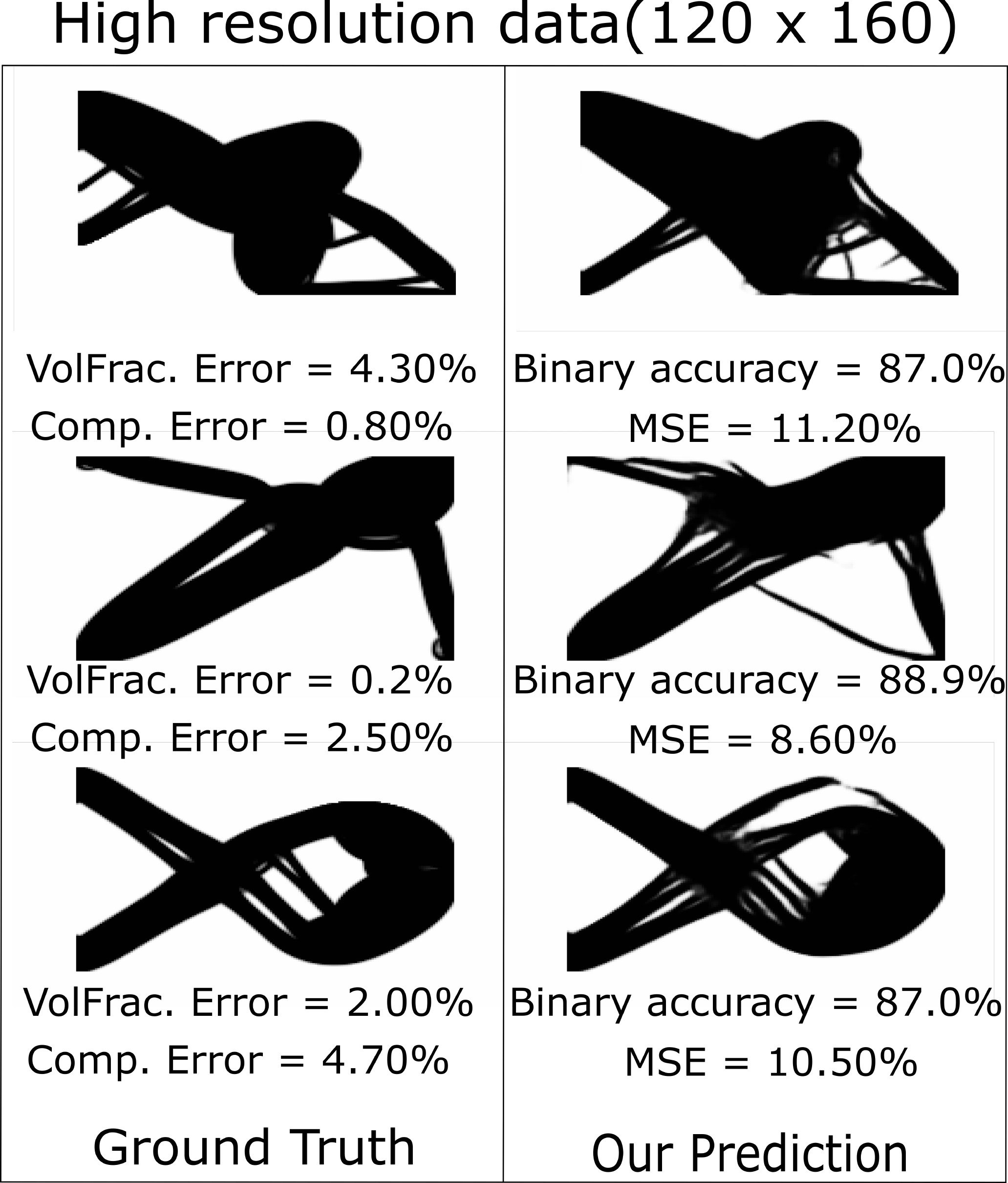}
      \caption{}
      \label{fig:2Dbeam-rect-unseen-2forcec}
    \end{subfigure}
    \begin{subfigure}[b]{.48\textwidth}
    \centering
      \includegraphics[width=.5\linewidth]{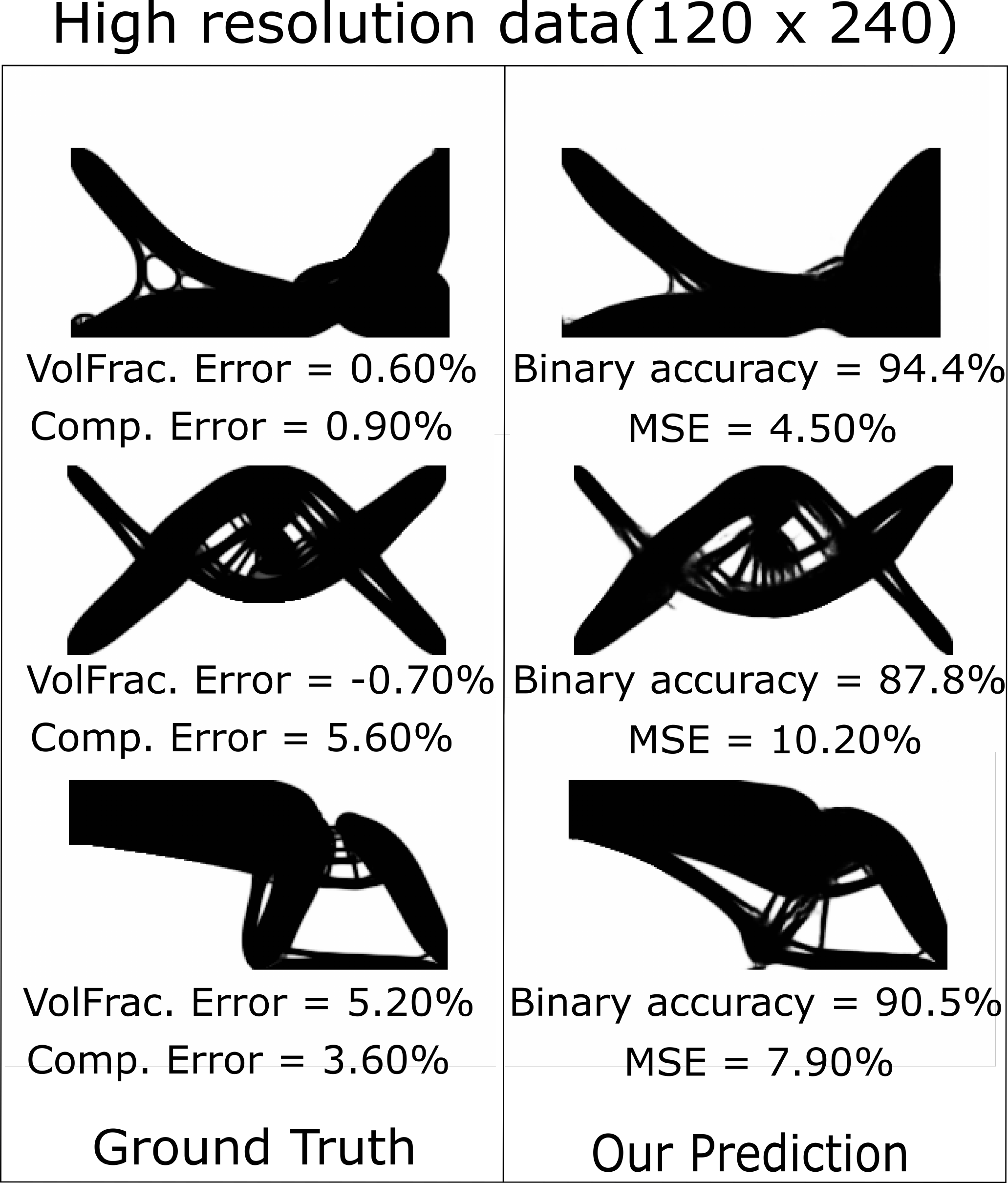}
      \caption{}
      \label{fig:2Dbeam-rect-unseen-2forced}
    \end{subfigure}
\hspace{-4.0cm}
    \begin{subfigure}[b]{.48\textwidth}
    \centering
      \includegraphics[width=.5\linewidth]{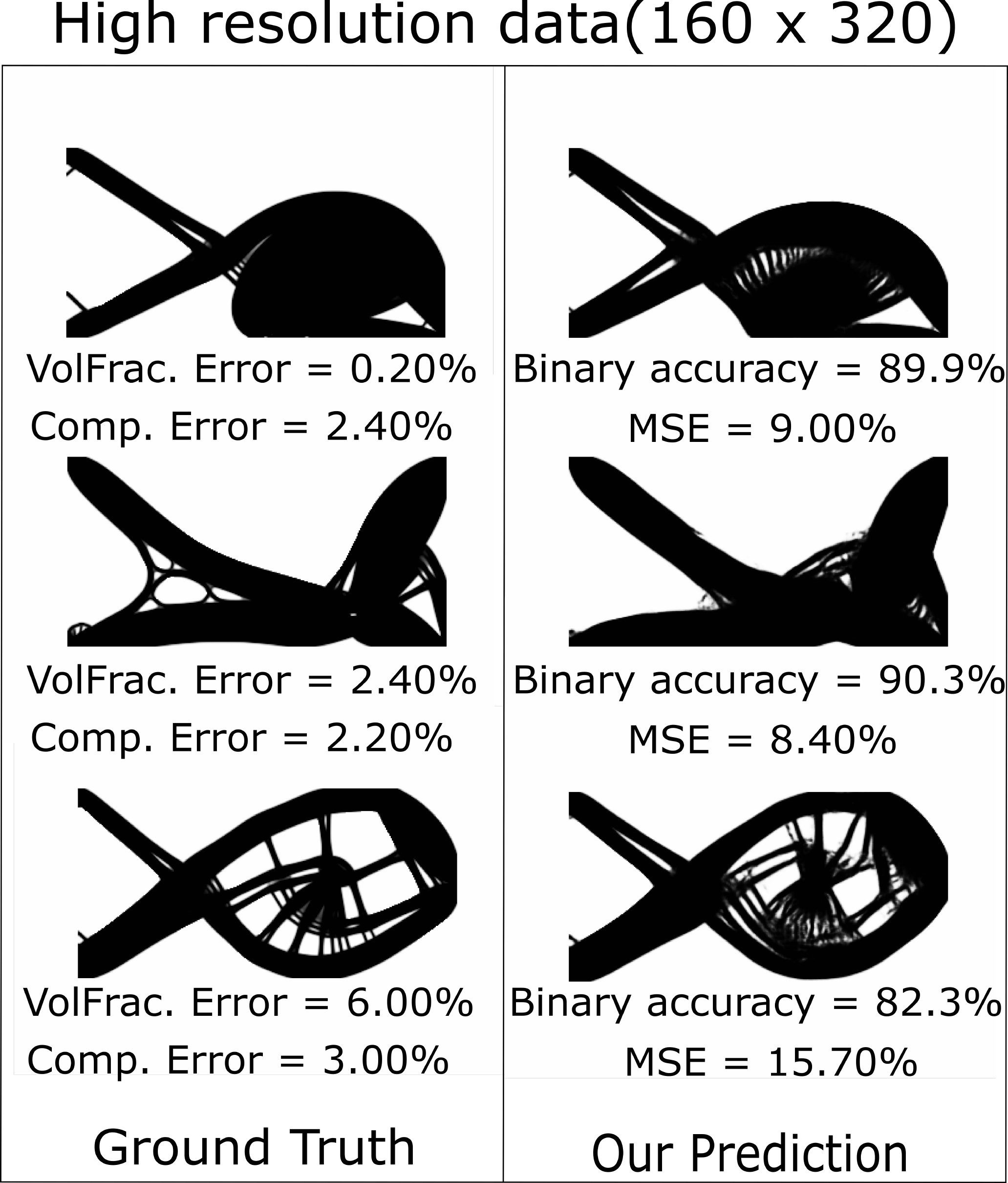}
      \caption{}
      \label{fig:2Dbeam-rect-unseen-2forcee}
    \end{subfigure}
\hspace{-4.0cm}
    \begin{subfigure}[b]{.48\textwidth}
    \centering
      \includegraphics[width=.5\linewidth]{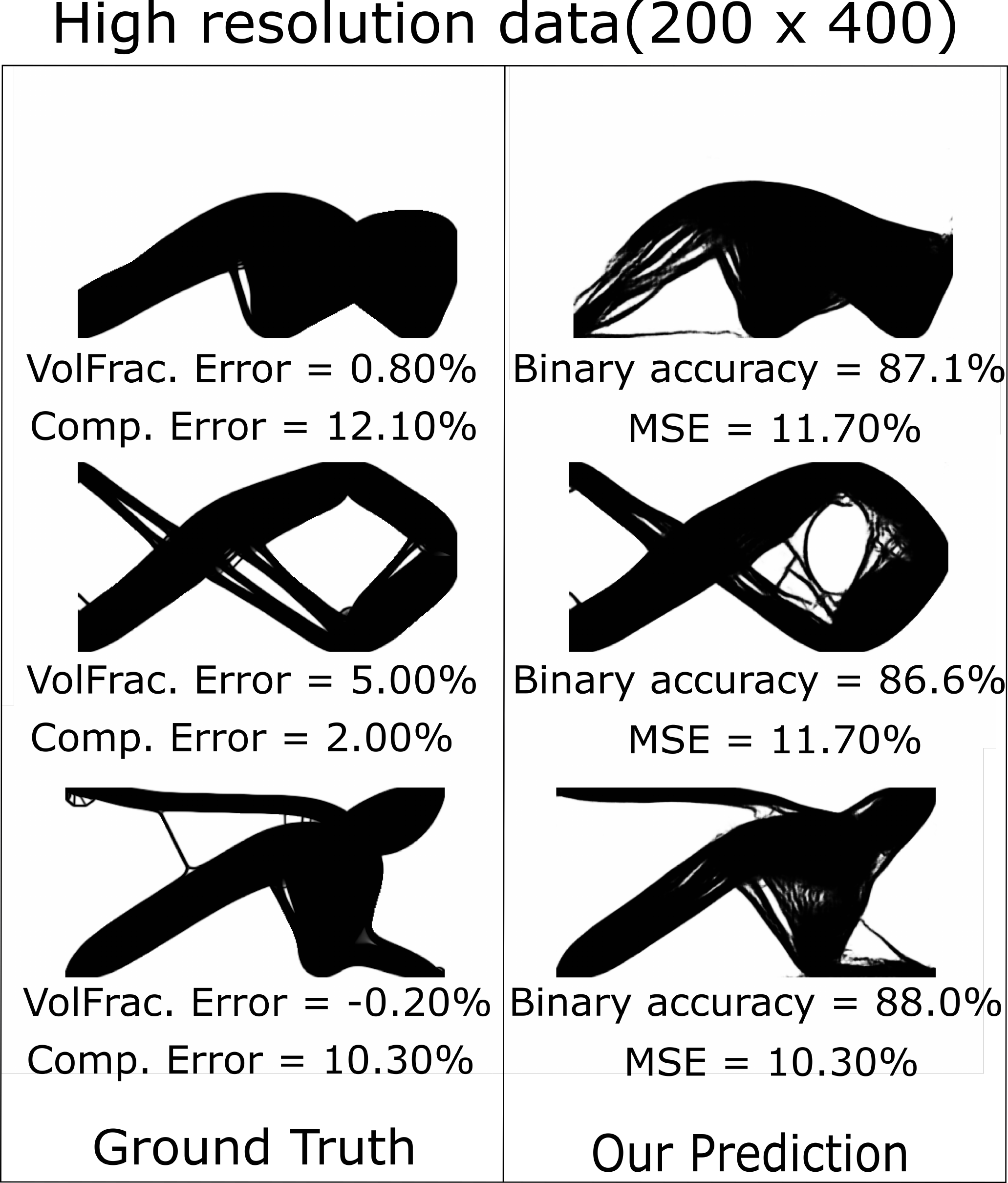}
      \caption{}
      \label{fig:2Dbeam-rect-unseen-2forcef}
    \end{subfigure}
\caption{Predictions vs. ground Truth for \textit{unseen} boundary conditions and \textit{two external forces}:  (a) the prediction of our source GAN alone; (b-f) optimal structures output by the fine tuned target model. The corresponding quality metrics are presented in Table \ref{tab:2Daccuracies-unseen-2force}.}
\label{fig:2Dbeam-rect-unseen-2force}

\end{figure*}

\begin{figure*}[th!]
\vspace{3pt}
\begin{center}
    \begin{tabular}{cc}
        \includegraphics[width=\columnwidth]{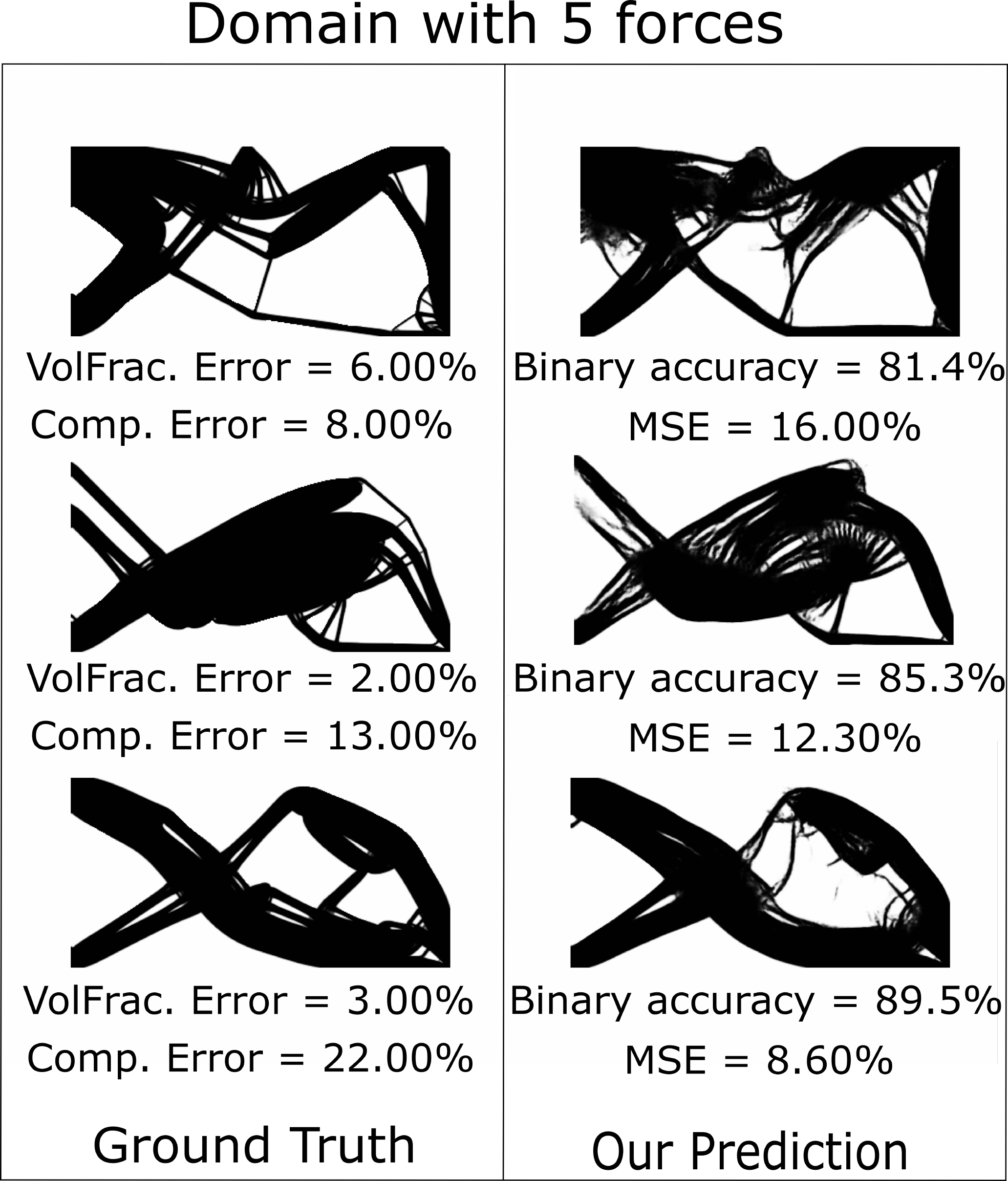}
        &
        \includegraphics[width=\columnwidth]{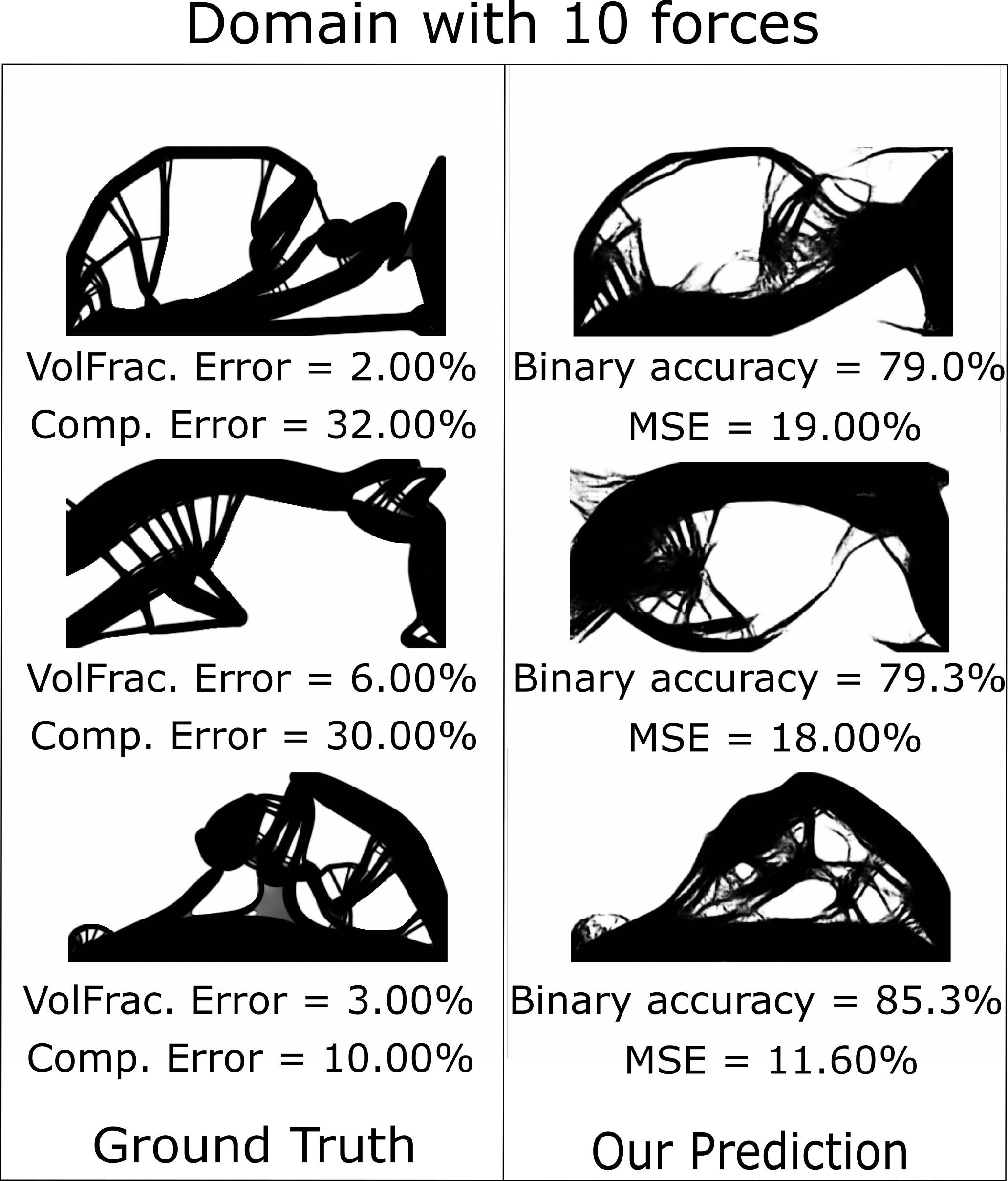}
        \\
        (a) & (b)\\
    \end{tabular}
    \\
\end{center}
\caption{Predictions vs. ground Truth for \textit{seen} boundary conditions and \textit{5} (a) and \textit{10} (b) external forces for a $200 \times 400$ domain resolution. } 
\label{fig:5-10 forces} \vspace{3pt} 
\end{figure*}

Figures \ref{fig:2Dbeam-rect}--\ref{fig:5-10 forces} show the side by side comparisons for all testing scenarios summarized in Table \ref{tab:scenarios}, and the corresponding evaluation data captured by the performance metrics described in section \ref{training:sec} was collected in tables \ref{tab:2Daccuraciesseen}--\ref{tab:2Daccuracies-unseen-2force}. The low resolution ($40 \times 80$) predictions in all these figures show the prediction of our source GANs and the comparison with the equivalent low resolution ground truth provided by SIMP.

The performance data shows that our network achieves a better prediction performance compared to the state-of-the-art deep learning-based methods. 

Specifically, for scenario \# 1 (\textit{seen} BCs with one randomly selected external force), the average MSE is 2.12\%, and the average binary accuracy is 96.72 with a 1\% average compliance error and a 2.5\% standard deviation. This implies that our network captures very well the changes in the topological patterns induced by a \textit{continuous} variation in the location, direction, and magnitude of the external force.

For scenario \# 2 (\textit{unseen} BCs with one randomly selected external force), the average MSE is  9\%, and the average binary accuracy is 89\%, with an average compliance error of 7.7\% and standard deviation of 11.7\%. Given the fact that the GAN was trained on only two different boundary conditions illustrated in \ref{fig:BCBeam} (a), (b), this test scenario indicates that the GANTL network does capture some high-level dependence of the topological patterns on the discrete boundary conditions. We note that the errors are primarily coming from incorrect predictions of the thin members of the ground truth, although this is not surprising given the fact that the network learned the topological patterns from boundary conditions that are dissimilar from those considered in this scenario.

The next two test scenarios use an unseen loading case, i.e., two randomly selected external forces and both seen and unseen boundary conditions, as described above. Recall that all training cases considered only one externally applied force. We observe that the performance metrics for both scenarios \# 3 \& 4 from Table \ref{tab:scenarios}, which include partially or totally unseen information, are very similar: average MSE -- 7\% vs. 11\%, binary accuracy -- 90\% vs 87\%, average compliance error -- 10.5\% vs 11.5\%, and standard deviation -- 16.2\% vs 15.7\%. 

In order to evaluate the performance of our predictions as the number of external forces increases, we included test cases with five and ten external forces for domains with a $200 \times 400$ resolution and boundary conditions displayed in Figure \ref{fig:BCBeam}(a) and (b). The side-by-side comparison between our predictions and the SIMP ground truths is shown in Figure \ref{fig:5-10 forces}. Although the average binary accuracy of the 250 test cases having five and ten external forces is around 84\% and 82\%, respectively, so only slightly lower than the data for scenarios \# 3 \& 4,  these figures indicate that as the number of external forces increases, the confidence of our network in predicting the thin members decreases. This behavior follows the known limitations of extrapolating approximations of non-linear functions, including the extrapolation of the neural networks learning outside of the support of the training distribution \cite{xu2021neural}. In some sense, this is conceptually similar with teaching a student the basic concepts of derivation and integration, then ask the student to solve differential equations. The student may solve simple ODEs, but would have a much harder time with complex ODEs or PDEs.

Collectively, the test scenarios \# 1-5 demonstrate that the proposed network is capable of high quality predictions of the overall optimal structure and clearly illustrate the generalization ability of our proposed GANTL. These test scenarios also indicate that the GANTL network has a better performance in predicting larger members of the optimal structure. At the same time, we observe that our network has a somewhat limited confidence in predicting the thinnest members of the optimal structure, and particularly for completely unseen boundary and loading conditions, and that this confidence decreases as the number of forces increases. This, however, is an expected behavior that is consistent with the observations from other machine learning-based predictors, and shows that the topological patterns learned by GANTL from the single external force used in training are changing as the number of external forces increases. Nevertheless, our results also suggest that constructing a training dataset that includes a reasonable sample of the practical loading cases used in practice would rapidly increase the quality of the network's predictions for design explorations. 

\begin{table*}[ht!]
    \centering
    \caption{MSE, binary accuracy, compliance error and standard deviation relative to SIMP for the structures shown in \ref{fig:2Dbeam-rect} with \textit{seen} BCs and one external force.}
    \begin{tabular}{lcSSSSS}
        \toprule
        Design Domain    & Resolution & {\shortstack{Number of \\ test cases}} & {MSE}  & {\shortstack{Binary \\Accuracy}} & {\shortstack{Compliance \\ Error}} & {\shortstack{Compliance \\ Error Std.}}\\
        \midrule
        {\shortstack{from Fig. \ref{fig:2Dbeam-recta} \\ (source network)}}  & 40 x 80 & 7473 & 0.48\% & 98.51\% & 0.32\% & 1.7\%     \\
        from Fig. \ref{fig:2Dbeam-rectb}  & 80 x 160  & 350  & 1.92\%  & 96.66\% & 1.03\% & 3.3\%  \\
        from Fig. \ref{fig:2Dbeam-rectc}   & 120 x 160 & 463  & 2.41\%  & 96.36\% & 1.08\% & 3.8\%  \\
        from Fig. \ref{fig:2Dbeam-rectd}   & 120 x 240 & 388  & 2.28\%  & 96.56\% & 0.98\% & 1.9\% \\
        from Fig. \ref{fig:2Dbeam-recte}  & 160 x 320 & 350  & 2.90\%  & 96.06\% & 1.14\% & 2.6\%    \\
        from Fig. \ref{fig:2Dbeam-rectf}  & 200 x 400 & 350  & 2.77\%  & 96.20\% & 1.45\% & 1.9\%  \\
        
  		\bottomrule
  		 		\textit{Average} &	&	&  2.12\% & 96.72\% & 1.00\% & 2.5\% \\
        \bottomrule
    \end{tabular}
    
    \label{tab:2Daccuraciesseen}
\end{table*}

\begin{table*}[ht!]
    \centering
    \caption{MSE, binary accuracy, compliance error and standard deviation relative to SIMP for the structures shown in \ref{fig:2Dbeam-rect-unseen-1force} with \textit{unseen} BCs and one external force.}
    \begin{tabular}{lcSSSSS}
        \toprule
        Design Domain    & Resolution & {\shortstack{Number of \\ test cases}} & {MSE}  & {\shortstack{Binary \\Accuracy}} & {\shortstack{Compliance \\ Error}} & {\shortstack{Compliance \\ Error Std.}}\\
        \midrule
        {\shortstack{from Fig. \ref{fig:2Dbeam-rect-unseen-1forcea} \\ (source network)}}  & 40 x 80 & 1000 & 8.24\% & 88.52\% & 8.31\% & 11.4\%     \\
        from Fig. \ref{fig:2Dbeam-rect-unseen-1forceb}  & 80 x 160  & 500  & 8.81\%  & 88.91\% & 6.65\% & 9.43\%  \\
        from Fig. \ref{fig:2Dbeam-rect-unseen-1forcec}   & 120 x 160 & 500  & 8.67\%  & 89.64\% & 8.09\% & 12.43\%  \\
        from Fig. \ref{fig:2Dbeam-rect-unseen-1forced}   & 120 x 240 & 500  & 8.92\%  & 89.32\% & 7.03\% & 11.92\% \\
        from Fig. \ref{fig:2Dbeam-rect-unseen-1forcee}  & 160 x 320 & 500  & 8.97\%  & 89.65\% & 6.91\% & 11.80\%    \\
        from Fig. \ref{fig:2Dbeam-rect-unseen-1forcef}  & 200 x 400 & 250  & 10.07\%  & 88.53\% & 9.47\% & 13.51\%  \\
        
  		\bottomrule
  		 		\textit{Average} &	&	&  8.94\% & 89.10\% & 7.74\% & 11.74\% \\
        \bottomrule
    \end{tabular}
    
    \label{tab:2Daccuracies-unseen-1force}
\end{table*}

\begin{table*}[ht!]
    \centering
    \caption{MSE, binary accuracy, compliance error and standard deviation relative to SIMP for the structures shown in \ref{fig:2Dbeam-rect-seen-2force} with \textit{seen} BCs and \textit{two external forces}.}
    \begin{tabular}{lcSSSSS}
        \toprule
        Design Domain    & Resolution & {\shortstack{Number of \\ test cases}} & {MSE}  & {\shortstack{Binary \\Accuracy}} & {\shortstack{Compliance \\ Error}} & {\shortstack{Compliance \\ Error Std.}}\\
        \midrule
         {\shortstack{from Fig. \ref{fig:2Dbeam-rect-seen-2forcea} \\ (source network)}}  & 40 x 80 & 1000 & 5.84\% & 90.98\% & 7.74\% & 13.2\%     \\
        from Fig. \ref{fig:2Dbeam-rect-seen-2forceb}  & 80 x 160  & 500  & 7.50\%  & 90.04\% & 16.12\% & 20.85\%  \\
        from Fig. \ref{fig:2Dbeam-rect-seen-2forcec}   & 120 x 160 & 500  & 8.91\%  & 90.01\% & 9.74\% & 16.88\%  \\
        from Fig. \ref{fig:2Dbeam-rect-seen-2forced}   & 120 x 240 & 500  & 7.67\%  & 90.49\% & 9.50\% & 15.30\% \\
        from Fig. \ref{fig:2Dbeam-rect-seen-2forcee}  & 160 x 320 & 500  & 8.32\%  & 90.13\% & 8.86\% & 14.62\%    \\
        from Fig. \ref{fig:2Dbeam-rect-seen-2forcef}  & 200 x 400 & 250  & 8.42\%  & 90.09\% & 10.92\% & 16.52\%  \\
        
  		\bottomrule
  		 		\textit{Average} &	&	&  7.77\% & 90.29\% & 10.48\% & 16.22\% \\
        \bottomrule
    \end{tabular}
    
    \label{tab:2Daccuracies-seen-2force}
\end{table*}


\begin{table*}[ht!]
    \centering
    \caption{MSE, binary accuracy, compliance error and standard deviation relative to SIMP for the structures shown in \ref{fig:2Dbeam-rect-unseen-2force} with \textit{unseen} BCs and \textit{two external forces}.}
    \begin{tabular}{lcSSSSS}
        \toprule
        Design Domain    & Resolution & {\shortstack{Number of \\ test cases}} & {MSE}  & {\shortstack{Binary \\Accuracy}} & {\shortstack{Compliance \\ Error}} & {\shortstack{Compliance \\ Error Std.}}\\
        \midrule
        {\shortstack{from Fig. \ref{fig:2Dbeam-rect-unseen-2forcea} \\ (source network)}}  & 40 x 80 & 1000 & 9.40\% & 87.95\% & 10.25\% & 14.1\%     \\
        from Fig. \ref{fig:2Dbeam-rect-unseen-2forceb}  & 80 x 160  & 500  & 10.60\%  & 86.89\% & 12.73\% & 16.98\%  \\
        from Fig. \ref{fig:2Dbeam-rect-unseen-2forcec}   & 120 x 160 & 500  & 10.81\%  & 87.20\% & 11.83\% & 16.61\%  \\
        from Fig. \ref{fig:2Dbeam-rect-unseen-2forced}   & 120 x 240 & 500  & 10.85\%  & 87.18\% & 11.21\% & 14.87\% \\
        from Fig. \ref{fig:2Dbeam-rect-unseen-2forcee}  & 160 x 320 & 500  & 11.68\%  & 86.66\% & 11.96\% & 16.11\%    \\
        from Fig. \ref{fig:2Dbeam-rect-unseen-2forcef}  & 200 x 400 & 250  & 11.86\%  & 86.47\% & 11.23\% & 15.81\%  \\
        
  		\bottomrule
  		 		\textit{Average} &	&	&  10.86\% & 87.05\% & 11.53\% & 15.74\% \\
        \bottomrule
    \end{tabular}
    
    \label{tab:2Daccuracies-unseen-2force}
\end{table*}


\subsection{A Topology-Aware Loss Function}

One common challenge of all published machine learning approaches in topology optimization, which exclusively use for now an image-based approach, is in predicting connected structural members. In fact, all these machine learning-based TO approaches simply fail to consider the topological connectivity of their predictions. Notably, even the best performing such methods attempt to improve the quality of the predictions by merely diversifying the training set. In the context of TO, this strategy increases the number of required training cases and, therefore, the associated computational cost of generating the dataset and of training the network, which may also impact the convergence of the training process.

Motivated by this observation, we propose a promising alternative for stimulating the network to produce connected structural members. Specifically, we construct and evaluate a custom loss function that is based on concepts from persistence homology \cite{edelsbrunner2008persistent, cohen2007stability}, and specifically on topological persistence diagrams. Informally, these persistence diagrams extract the `birth' and the `death' of the topological features (e.g., connected components) in the structure as the structure evolves according to predefined filters. By correlating the persistence diagram of the ground truth with that of the prediction and by including this correlation in the custom loss function, we push the network to make predictions that are topologically similar to the ground truth.

For these experiments, we used the bottleneck distance between persistence diagrams, which can be thought of as  the shortest `distance' $b$ for which there exist an optimal matching between points of two given persistence diagrams such that any pair of matched points are at distance at most $b$ \cite{edelsbrunner2008persistent,efrat2001geometry}.

We add the custom loss function to the binary cross-entropy and use the total loss as the GAN loss function:
\begin{equation}
    L(y_{t},y_{p}) = L_{bce}(y_{t},y_{p}) + \lambda L_{topology}(y_{t},y_{p})
\end{equation}
where $y_{t}$ is the ground truth structure, $y_{p}$ is the predicted structure, $L_{bce}$ is the binary cross-entropy, $L_{topology}$ is the bottleneck distance between the persistent diagrams, and $\lambda$ controls the weight of the topological loss which can be in (0,1]. In our experiments, we chose $\lambda$ to be 0.1. We extract the persistence diagrams of structures by using Ripser \cite{ctralie2018ripser} and the bottleneck distance is computed with the GUDHI library \cite{maria2014gudhi}.

We fine-tuned the target GAN for our highest resolution (200 x 400) domains, which also produced predictions with the highest number of disconnected members in the experiments detailed above, with the new topology-aware loss function.  Prior to displaying the predicted structures, we round up or down the density values to the closest integer, either 0 or 1. We tested the performance of the proposed topology-aware loss function for both seen and unseen displacement boundary conditions.

\begin{figure*}[th!]
\vspace{3pt}
\begin{center}
    \begin{tabular}{cc}
        \includegraphics[width=\columnwidth]{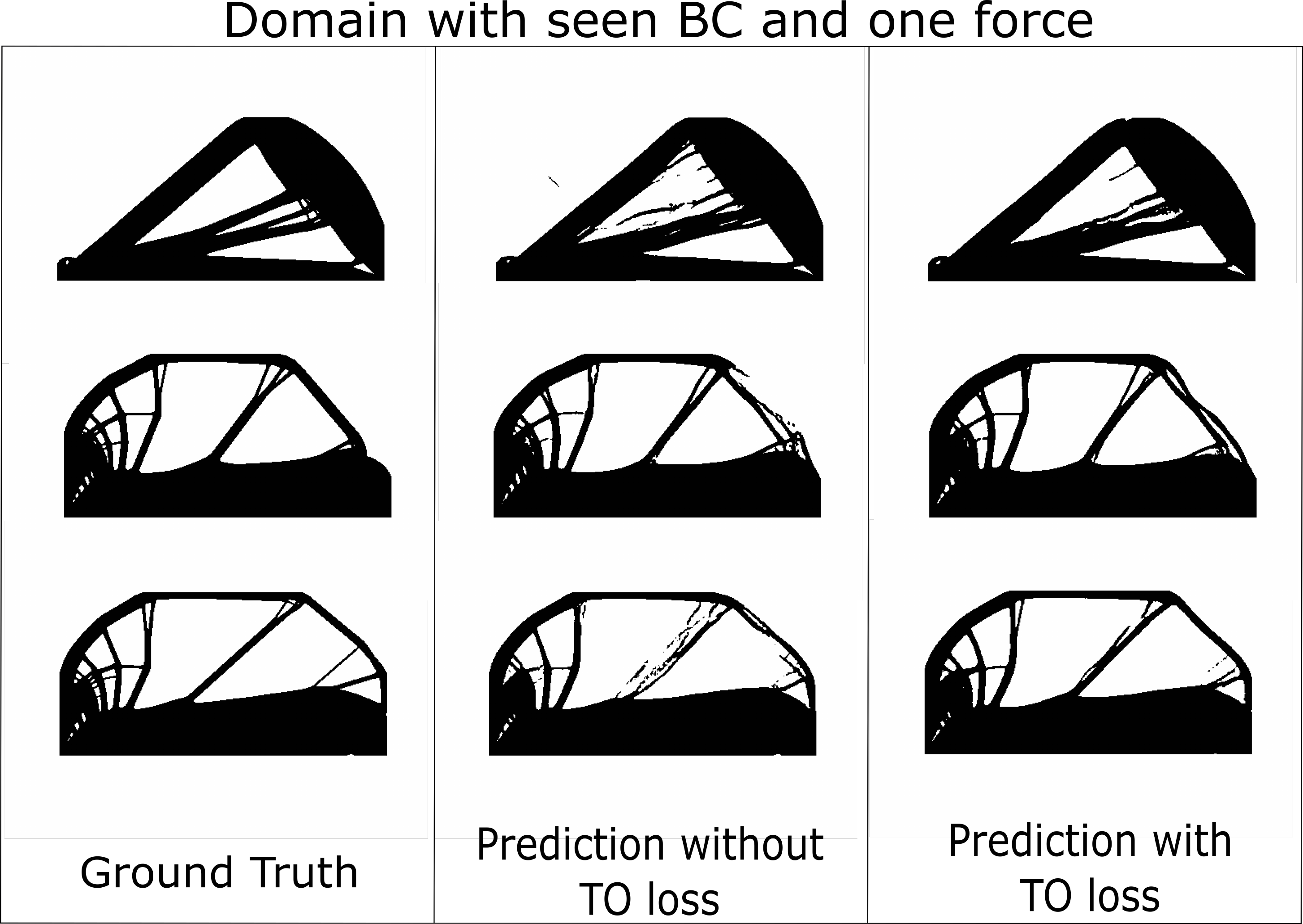}
        \hspace{15pt}
        &
        \includegraphics[width=\columnwidth]{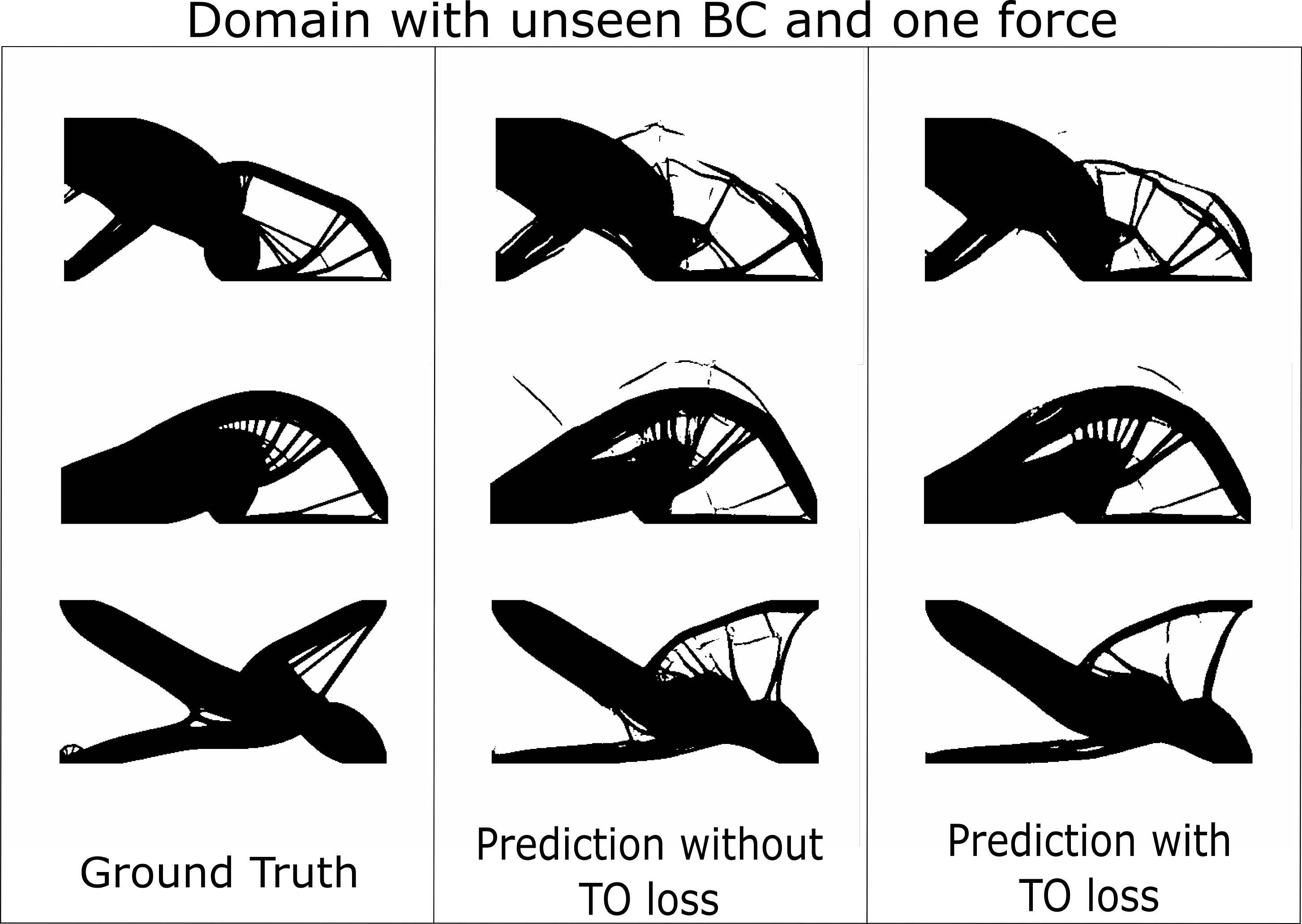}
        \hspace{15pt}
        \\
        (a) & (b)\\
    \end{tabular}
    \\
    \begin{tabular}{cc}
        \includegraphics[width=1\columnwidth]{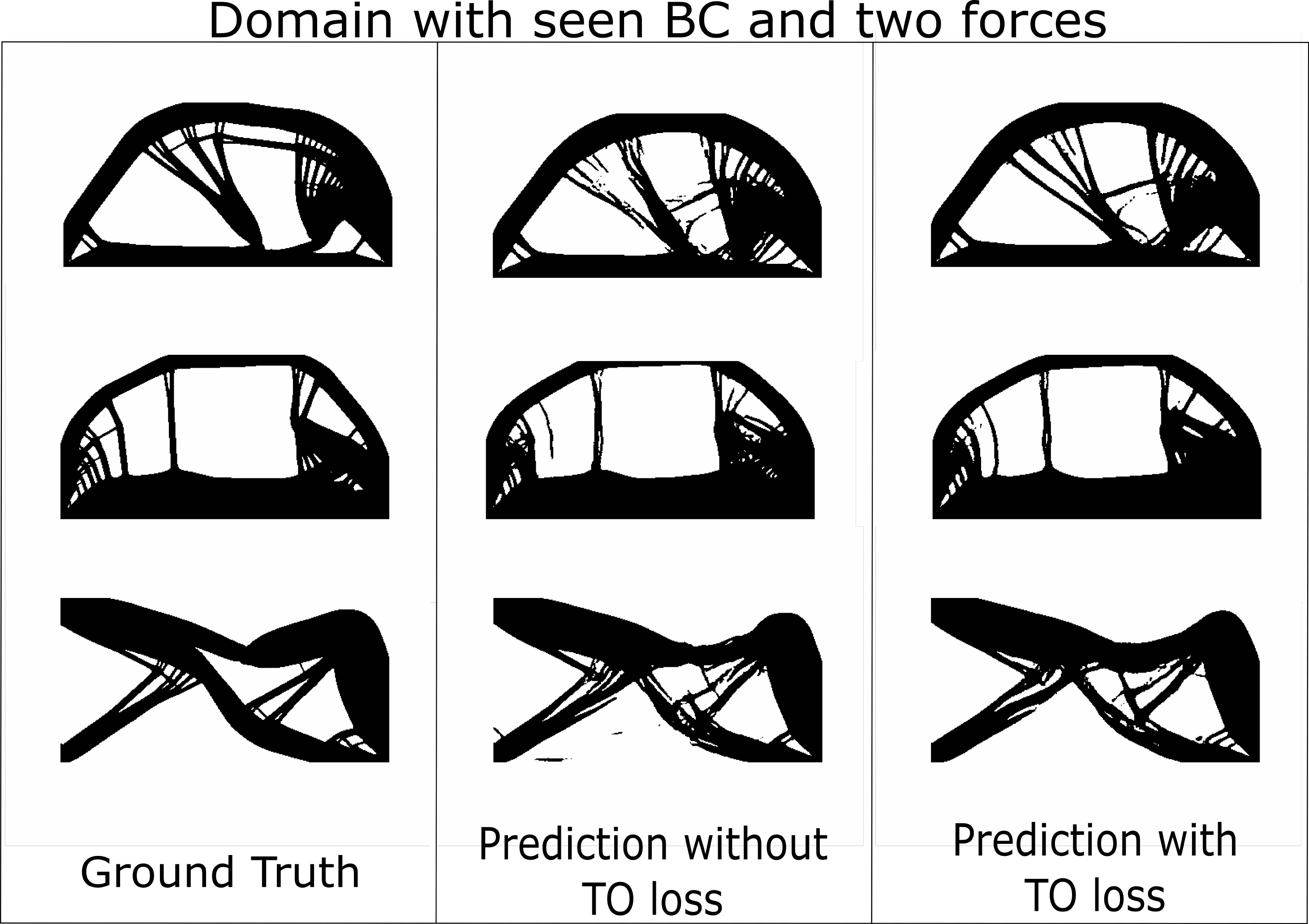}
        \hspace{15pt}
        &
        \includegraphics[width=1\columnwidth]{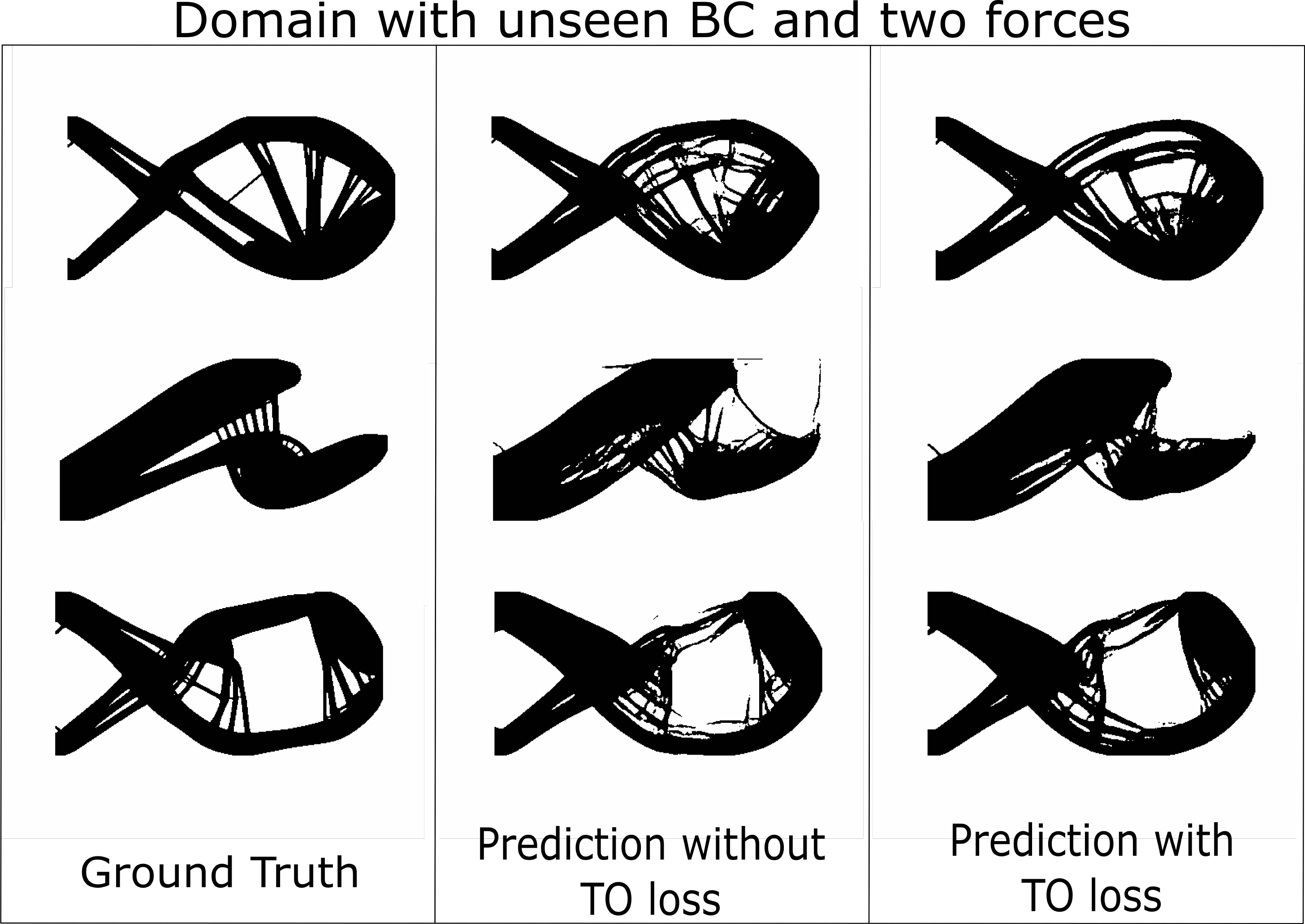}
        \hspace{15pt}
        \\
        (c) & (d) \\
    \end{tabular}
    \\
    \begin{tabular}{cc}
        \includegraphics[width=1\columnwidth]{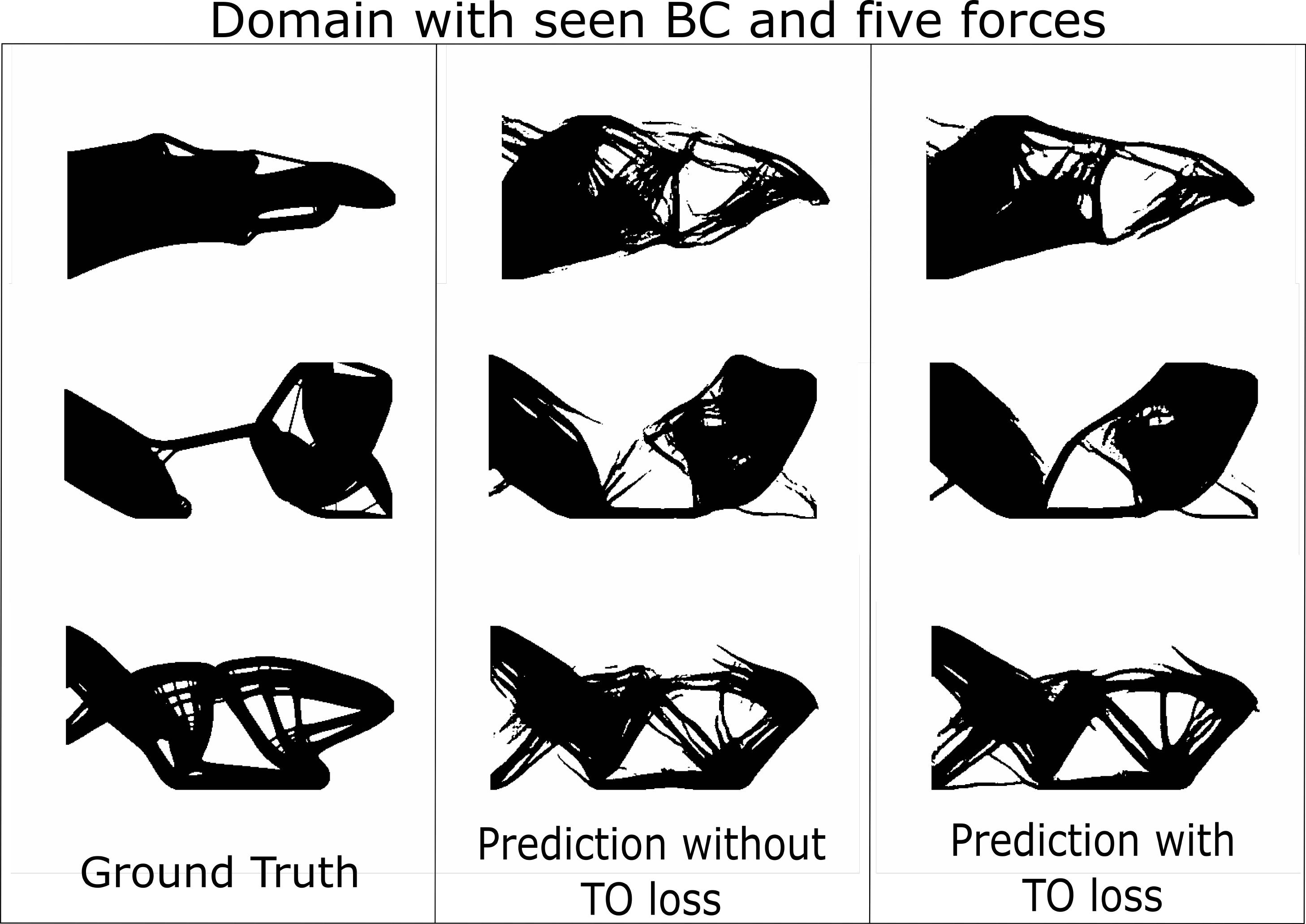}
        \hspace{15pt}
        &
        \includegraphics[width=1\columnwidth]{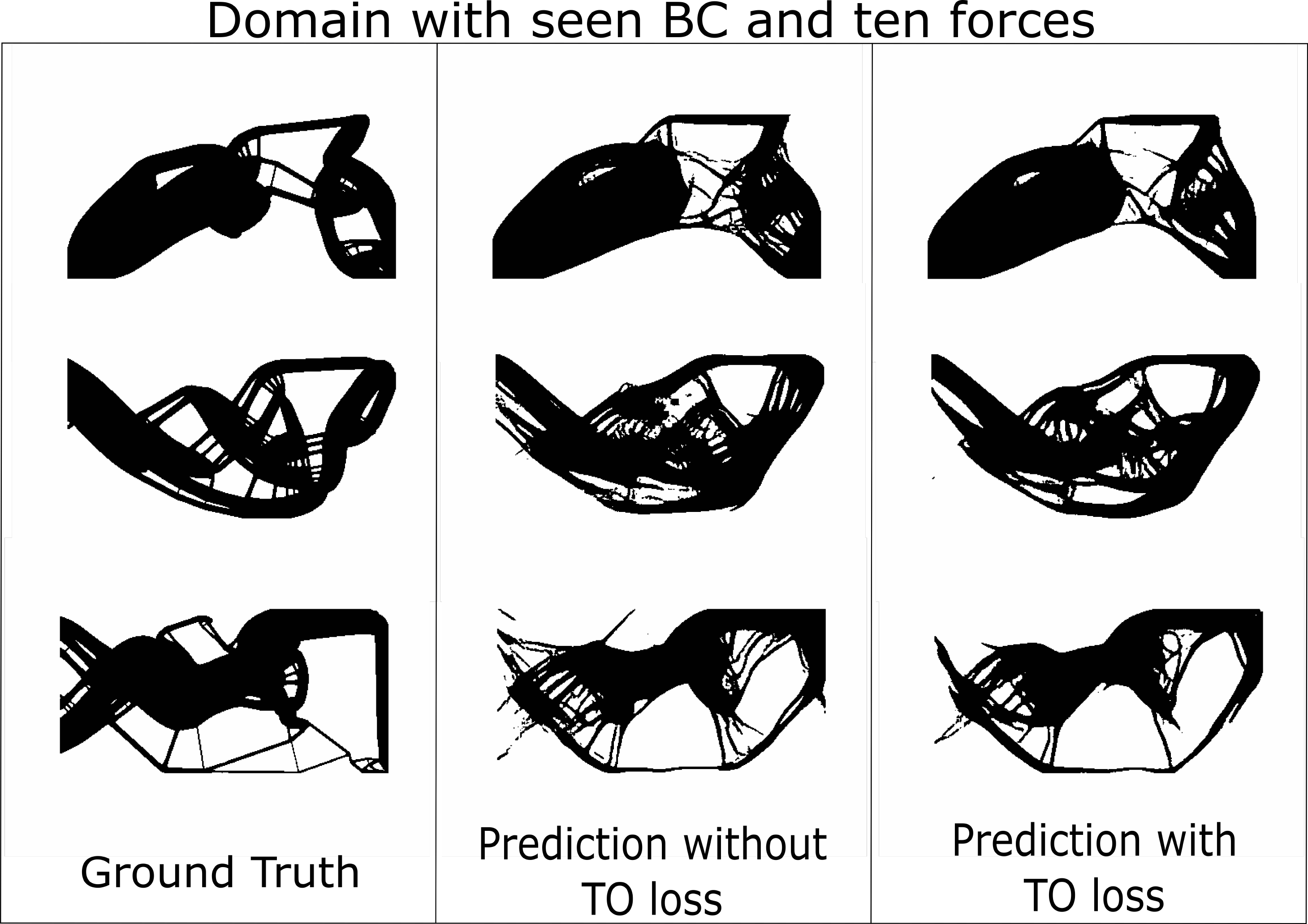}
        \hspace{15pt}
        \\
        (e) & (f) \\
    \end{tabular}
    \\
\end{center}
\caption{Comparison between ground truth, and the predictions using standard cross entropy loss (second column in each panel) and those using the proposed topology-aware loss function based on the bottleneck distance between the corresponding persistence diagrams (third column in each panel). (a) seen BCs, 1 external force; (b) unseen BCs and one external force; (c) seen BCs and 2 external forces; (d) unseen BCs and 2 external forces; (e) seen BCs and 5 external forces; (f) seen BCs and 10 external forces.} 
\label{fig:newloss} \vspace{3pt} 
\end{figure*}

As shown in Figure \ref{fig:newloss},  the proposed topology-aware loss function can significantly reduce the number of disconnected members for both seen and unseen boundary conditions, and also eliminates many small holes in our predictions.


\section{Conclusions}\label{conclusions:sec}

We proposed a highly efficient and accurate generative design tool that combines the power of generative adversarial networks with the efficiency provided by the knowledge reuse of transfer learning. The proposed method uses 11,000 low resolution cases to train a source generator, whose knowledge is transferred to a target network. Consequently, the training (fine-tuning) of the target network for high-resolution domains needs a much smaller dataset (1,500 cases).   The fact that we transferred the knowledge learned from the low resolution to the high resolution cases allowed us to significantly reduce the number of high resolution cases without losing the accuracy of the predictions. We presented a discussion of the associated computational costs of generating the training cases in \cite{behzadi2021real} so we do not repeat it here. 

We produced numerous examples to show that the proposed GANTL approach can generate highly accurate predictions for seen boundary conditions with a relatively small performance decrease for domains that use unseen boundary conditions that were not part of the training data. In addition, while all training cases had one single external force applied to the domain, our test data also included external loading with 2, 5, and 10 external loads. 

An analysis of these examples points to the fact that the confidence of our network in predicting the thin members decreases as the number of forces increases. This behavior is consistent with the behavior observed by all image-based machine learning algorithms, including all known approaches that use these algorithms in topology optimization, and with the limitations of extrapolating approximations of non-linear functions outside of their sampled domain. Specifically, our examples illustrate that the topological patterns learned by GANTL from training data using a single external force are changing as the number of external forces increases. 

Motivated by these observations, we also proposed a novel and promising alternative for stimulating the connectedness of the predictions produced by the network. Specifically, we constructed a custom loss function that correlates the persistence diagram of the ground truth with that of the prediction. We showed that this novel topology-aware loss function adequately promotes those predictions that are topologically similar with the topology of the ground truth, and results in predictions with a significant improvement in terms of the connectedness of the predicted structural members. We posit here that enforcing topological properties of the optimal structures predicted by GANTL can be achieved by developing specialized loss functions that integrate topological measures. However, a detailed investigation of these aspects is outside the scope of this paper. 

Training GANs can be notoriously difficult, as discussed above. Our network faced the same challenges,  which were addressed by employing a binary cross-entropy (BCE) with the logistic unit (logits) instead of using BCE alone. This modification helps the discriminator avoid loss values that are near zero and consequently improve the stability of the GAN.  

One key limitation in promoting predictions with topological properties that are similar to those of the ground truth structures is the added computational overhead of the topology-aware loss function. To compute this loss function, we had to compute the persistence diagram of all the ground truth structures and of the corresponding predicted structures, as well as their pairwise bottleneck distance. This, in turn, significantly increased the training time of our GAN model, although it was computed only once for the training set.

As we demonstrated in  \cite{behzadi2021real}, transfer learning not only significantly reduces the size of the dataset required to train the networks but improves the generalization ability of the deep learning models as long as the target network is fine-tuned with domains and boundary conditions that are not part the training dataset of the source model. Importantly, by combining the generative power of GANs with the knowledge reuse of transfer learning, we train both the source and target networks only once and therefore eliminate the need to fine-tune the target network every time we want to explore new boundary conditions. This is a critical feature of the proposed GANTL that significantly reduces the training time and directly improves the practicality of the approach for design space explorations.

To the best of our knowledge, this is the first attempt to combine transfer learning with generative adversarial networks in the area of topology optimization. Furthermore, our proposed topology-aware loss function pioneers further integrations of powerful topological measures into the TO predictions made by machine learning algorithms, which will eventually lead to practical, powerful, and interactive design exploration tools with topology optimization.

%

%
% ...

\bibliographystyle{asmems4}

\bibliography{asme2e}



\end{document}